\documentclass[10pt,conference]{IEEEtran}
\IEEEoverridecommandlockouts
\usepackage{cite}
\usepackage{amsmath,amssymb,amsfonts}
\usepackage{algorithmic}
\usepackage{graphicx}
\usepackage{textcomp}
\usepackage{xcolor}
\usepackage{url}
\usepackage{subfigure}
\usepackage{multirow}
\usepackage{enumitem}
\usepackage{booktabs}
\usepackage{makecell}
\usepackage[ruled,linesnumbered]{algorithm2e}
\newtheorem{definition}{Definition}
\newtheorem{exmp}{Example}

\def\BibTeX{{\rm B\kern-.05em{\sc i\kern-.025em b}\kern-.08em
    T\kern-.1667em\lower.7ex\hbox{E}\kern-.125emX}}
\begin{document}

\title{Realistic Safety-critical Scenarios Search for Autonomous Driving System via Behavior Tree}


\author{Ping Zhang$^1$,
	Lingfeng Ming$^{1,*}$,
	Tingyi Yuan$^2$,
	Cong Qiu$^1$,
	Yang Li$^1$,
    Xinhua Hui$^1$,
	Zhiquan Zhang$^1$,
    Chao Huang$^1$\\
    $^1$Alibaba Group, Hangzhou, China\\
    $^2$Xi'an Jiaotong University, Xian, China\\
    \IEEEcompsocitemizethanks{
		\IEEEcompsocthanksitem $^{*}$Lingfeng Ming is the corresponding author.}
		
		
		
}

\maketitle

\begin{abstract}

The simulation-based testing of Autonomous Driving Systems (ADSs) has gained significant attention. 
However, current approaches often fall short of accurately assessing ADSs for two reasons: over-reliance on expert knowledge and the utilization of simplistic evaluation metrics. 
That leads to discrepancies between simulated scenarios and naturalistic driving environments. 
To address this, we propose the Matrix-Fuzzer, a behavior tree-based testing framework, to automatically generate realistic safety-critical test scenarios. 
Our approach involves the $log2BT$ method, which abstracts logged road-users' trajectories to behavior sequences. 
Furthermore, we vary the properties of behaviors from real-world driving distributions and then use an adaptive algorithm to explore the input space. 
Meanwhile, we design a general evaluation engine that guides the algorithm toward critical areas, thus reducing the generation of invalid scenarios. 
Our approach is demonstrated in our Matrix Simulator. 
The experimental results show that: 
(1) Our $log2BT$ achieves satisfactory trajectory reconstructions. 
(2) Our approach is able to find the most types of safety-critical scenarios, but only generating around 30\% of the total scenarios compared with the baseline algorithm. 
Specifically, it improves the ratio of the critical violations to total scenarios and the ratio of the types to total scenarios by at least 10x and 5x, respectively, 
while reducing the ratio of the invalid scenarios to total scenarios by at least 58\% in two case studies.
\end{abstract}




\begin{IEEEkeywords}
Search-based Software Engineering, Behavior Tree, Safety-Critical scenario, Autonomous Driving System
\end{IEEEkeywords}

\section{Introduction}\label{sec:introduction}
\noindent
The potential for autonomous vehicles (AVs) to enhance driving safety over human drivers and alleviate traffic congestion has garnered considerable interest from industry and academia. 
These attentions have resulted in the brisk advancement of Automated Driving Systems (ADSs). Nevertheless, the safety performance of AVs and the subsequent paucity of public trust have impeded their widespread production and application. 
Consequently, the validation of ADSs has become critical.

Safety validation is a key process to improve the safety performance of AVs. 
A widely used validation method is on-road testing, as it directly reflects the performance of tested ADS in a real-world environment. 
However, due to the rarity of safety-critical events, millions of miles need to be tested in the naturalistic driving environment (NDE)\cite{feng2023dense}. 
Besides, it's impossible to evaluate all the corner cases. 
An alternative is to challenge ADSs in computer simulation. 
Simulation-based testing prevails over on-road testing in terms of consistency, scalability, and safety.
In recent years, many methods \cite{DBLP:conf/ivs/AlthoffL18,DBLP:journals/tits/FengFSZL22,DBLP:journals/tcps/KimMS20} have been proposed to construct virtual simulation scenarios to evaluate ADSs.


For the convenience of presentation, we follow the definition of "scenario" with three abstraction levels \cite{DBLP:conf/ivs/MenzelBM18,DBLP:conf/itsc/UlbrichMRSM15}: A \textbf{functional
scenario ($\mathcal{FS}$)} is a qualitative description (e.g., the participant vehicle cruises at a low speed), which derives a \textbf{logical scenario ($\mathcal{LS}$)} with property variables (e.g., the speed of the vehicle is between 5m/s and 10m/s). 
A \textbf{concrete test scenario ($\mathcal{CTS}$)} is a vector of exact variable values (e.g., the vehicle cruises at 6m/s).
Existing scenario-based approaches \cite{DBLP:conf/itsc/FremontKPSABWLL20,DBLP:conf/issre/LiLJTSHKI20,DBLP:conf/ivs/MenzelBISM19,DBLP:conf/ivs/MenzelBM18} aim to find all safety-critical $\mathcal{CTS}s$ in an $\mathcal{LS}$.

\begin{figure}[tbp]
	\centering
	\includegraphics[width=.48\textwidth]{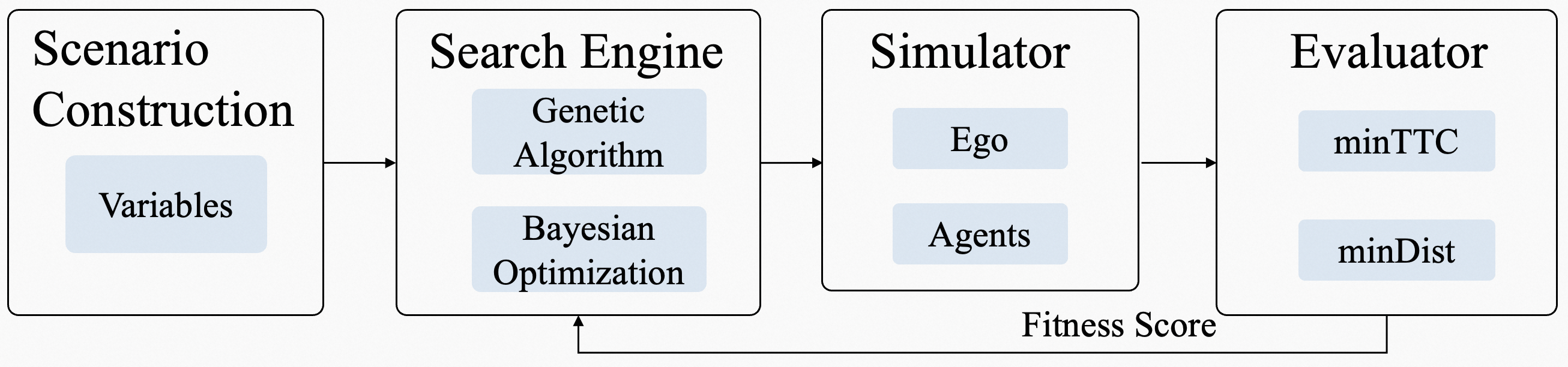}
	\caption{The general workflow of scenario-based ADS testing in simulators.}
	\label{fig:base_framework}
\end{figure}

We summarize the general workflow of the ADS validation in simulators. 
As shown in Fig. \ref{fig:base_framework}, it consists of four major components: 
(1) Scenario Construction, which aims to construct an $\mathcal{LS}$ with variables. (2) Search Engine, which adopts an optimization algorithm and samples the next vector of variables to generate a new $\mathcal{CTS}$. 
(3) Simulator, which simulates the traffic corresponding with the input $\mathcal{CTS}$. (4) Evaluator, which computes the fitness score and feeds it back to the search engine for the next iteration.

However, these approaches mainly face the three challenges:
\begin{enumerate}
	\item \textbf{High-dimensional space}: The driving environments are spatial-temporally complex, 
	and the high-dimensional variables are needed to define such an environment.
	These approaches \cite{DBLP:conf/ivs/MenzelBM18,DBLP:conf/ijcai/Wachi19,DBLP:conf/icra/Abeysirigoonawardena19,feng2023dense,norden2019efficient,DBLP:conf/icra/KuuttiFB20,DBLP:journals/tits/ChenCWL22} parameterize the states of traffic participants\footnote{Agents and participants are equal in this paper.} (e.g., vehicle/bicycle/human) at every step, suffering from the curse of dimension. 
	As the input space grows, the computational complexity grows exponentially, making it much more difficult to find the rare safety-critical $\mathcal{CTS}s$.
  
  	\item \textbf{Non-NDE scenarios}: 
	On one hand, one prevalent way to exploit real-world scenarios is to replay the pre-recorded scenario directly to challenge ADSs. 
	Naturalistic as their original layout and trajectories are, these scenarios become invalid when ADSs act differently, due to the absence of responsive agents.
	CRISCO \cite{DBLP:conf/kbse/TianWYJ00LY22} challenges ADSs by assigning participants to perform influential behaviors mined from NDE. 
	However, it's not adequate to change behavior patterns without parameterizing properties of behaviors.
	On the other hand, the scenario-based studies \cite{DBLP:conf/itsc/FremontKPSABWLL20,DBLP:conf/issre/LiLJTSHKI20,DBLP:conf/ivs/MenzelBISM19,DBLP:conf/ivs/MenzelBM18} construct $\mathcal{LS}s$ via expert knowledge. 
	Inevitably, these $\mathcal{LS}s$ deviate from NDE to some extent, resulting in the generated $\mathcal{CTS}s$ different from the real-world driving distributions.

	\item \textbf{Incorrect Evaluation}: A $\mathcal{CTS}$ is considered critical if a safety measure is below (over) a user-specified threshold. 
	However, there is no adequate evaluation. 
	Many studies \cite{feng2023dense,DBLP:conf/itsc/GangopadhyayKDD19,DBLP:conf/issre/LiLJTSHKI20,DBLP:conf/kbse/TianWYJ00LY22,DBLP:conf/amcc/ZhouR18} apply $minTTC$ (Time To Collision) or $minDist$ (distance between the ego and participants). 
	These studies ignore the reasonability of the participants' behaviors, misleading the search algorithm to a wrong direction and thus generating many unrealistic $\mathcal{CTS}s$.
	In particular, the reckless road users lead to unavoidable accidents even when ADSs are functioning properly. 
	The reckless behaviors go against the function testing of ADSs.
\end{enumerate}

To solve the above challenges, we propose Matrix-Fuzzer, a behavior tree-based testing framework, to search for realistic safety-critical scenarios for ADS testing. 
We aim to generate scenarios where an AV can run into safety violations by fuzzing the traffic participants' initial states, behavioral trigger conditions, and driving behaviors. 
We transform the variety of variables into a parameter optimization problem. 
First, we propose $log2BT$ that converts raw logged trajectories into high-level behavior sequences. 
Then, we adopt real-world driving distributions to vary the properties for the behavior tree-based $\mathcal{LS}$ construction. 
Furthermore, we design an adequate evaluation engine to compute the fitness scores of test scenarios and apply an adaptive algorithm for the input space to solve this optimization problem. 
Finally, we classify the generated safety-critical $\mathcal{CTS}s$ and analyze the root causes. 

Overall, the major contributions are summarized as follows:
\begin{itemize}
	\item We propose Matrix-Fuzzer, a behavior tree-based testing framework, to search realistic safety-critical scenarios for ADS. 
	It is more comprehensible and editable to describe driving environments with high-level behaviors, which greatly reduces the number of variables.

  	\item We propose $log2BT$, which converts the raw logged trajectories into the high-level behavior sequences. 
	The variable spaces of behavior tree-based $\mathcal{LS}s$ are determined from realistic driving distributions.
  	
	\item We design an adequate evaluation engine to evaluate the rationality of participants and the ego, which guides the adaptive algorithm to avoid generating invalid scenarios.
	
	\item We classify the safety-critical scenarios reported by Matrix-Fuzzer in our Matrix Simulator and analyze the root causes of the violations. 
	The experimental results show that our approach outperforms the baseline methods. 
	Note that, Matrix-Fuzzer has been applied in our platform and provides daily massive simulation services.
\end{itemize}


\section{Related Work}\label{sec:related}

\subsection{Variable For Scenario Search}
\noindent
In scenario-based ADS testing, any property of driving environments could be described as a variable to construct different scenarios.
The variables are summarized in the survey \cite{DBLP:journals/corr/abs-2112-00964} according to the scenario layer model.
Existing studies use a set of variables. Typical variables include road curvature \cite{DBLP:conf/kbse/AbdessalemNBS16}, type of road and number of lanes \cite{DBLP:conf/itsc/NitscheWGT18,DBLP:journals/corr/abs-2109-06126,DBLP:conf/icra/TangZWLSHW21,DBLP:conf/ivs/TangZLSW21},
traffic infrastructures (e.g., traffic signs) \cite{DBLP:conf/kbse/AbdessalemPNBS18}, the initial state (e.g., pose, speed, heading) of the ego and participants \cite{DBLP:conf/kbse/AbdessalemPNBS18,DBLP:conf/hybrid/TuncaliFIK18,DBLP:conf/pts/KluckZWN19,DBLP:conf/qrs/KluckZWN19,DBLP:journals/tiv/TuncaliFPIK20,DBLP:journals/infsof/LiTW20},
the trajectories of the ego and participants \cite{DBLP:conf/ijcai/Wachi19,DBLP:conf/icra/Abeysirigoonawardena19,DBLP:conf/icra/KuuttiFB20,DBLP:journals/tits/ChenCWL22}, and the environmental conditions like weather and lighting \cite{DBLP:journals/corr/abs-2109-06404,DBLP:journals/corr/abs-2109-06126,DBLP:conf/kbse/AbdessalemNBS16}.
Among them, the trajectory generalization is critical. 
However, controlling object's state at every time step suffers from high-dimensional space, which limits its application.

On the one hand, an alternative is to abstract the trajectories into high-level behaviors to shrink the input space.
These studies \cite{DBLP:conf/ivs/BusslerHPS20,DBLP:journals/ral/DingCLEZ21,DBLP:journals/corr/abs-2109-06126,DBLP:journals/corr/abs-2109-06404,DBLP:conf/ivs/GhodsiHFTTKGA21,DBLP:conf/iros/DingCXZ20} adopt
the variables of conditions and behaviors to control objects. But there is no correlation between these behaviors, suffering from scalability for complex traffic environments.
Behavior tree ($\mathcal{BT}$) \cite{DBLP:journals/corr/abs-1709-00084} is a very efficient way of creating complex systems that are both
modular and reactive.
Unreal Engine 4 (UE4)\cite{unrealengine} applies $\mathcal{BT}s$ to create behaviors for non-player characters (NPC).
RoadRunner Scenario \cite{roadrunner}, a popular 3D scenario editor, adopts $\mathcal{BT}s$ to edit participants' behaviors.
But there are some drawbacks in RoadRunner Scenario, such as not supporting
non-motor vehicles, being unable to control traffic signal, and requiring external programs to achieve scenario generalization.
To be more comprehensible and editable, the proposed Matrix-Fuzzer also applies $\mathcal{BT}$ to describe the driving environments.

On the other hand, it is more consistent with requirements to generate the test scenarios from NDE.
However, the above studies edit scenarios and set variable ranges by artificial expertise.
Different from knowledge-based scenarios \cite{DBLP:conf/itsc/FremontKPSABWLL20,DBLP:conf/issre/LiLJTSHKI20,DBLP:conf/ivs/MenzelBISM19,DBLP:conf/ivs/MenzelBM18}, 
CRISCO \cite{DBLP:conf/kbse/TianWYJ00LY22} mines influential behavior patterns from traffic trajectories and then generates $\mathcal{CTS}s$ varying behaviors patterns. 
Specifically, it first identifies collision-related trajectories, then extracts the high level behaviors by a simple segment mechanism, and generates test scenarios by differing behavior patterns from four aspects.
To generalize the behavior patterns of participants rather than from pre-defined aspects, we design $log2BT$ to abstract traffic trajectories into high-level behavior sequences described by behavior trees, and vary their properties to generate $\mathcal{CTS}s$.

\subsection{Concrete Test Scenarios Search}
\noindent
Given an $\mathcal{LS}$ containing variables, the goal of simulation-based ADS testing is to quickly find all safety-critical $\mathcal{CTS}s$ in the search space. 
Random search and grid search are commonly used, but they are inefficient in high-dimensional space.



Therefore, heuristic search algorithms are popular in safety-critical scenario search.
Bayesian optimization (BO) \cite{DBLP:journals/pieee/ShahriariSWAF16} is a classic machine-learning method for solving optimization problems.
The study \cite{DBLP:conf/itsc/GangopadhyayKDD19} adopts BO to search critical scenarios in human crossing the road environment, while BO is used to optimize the policy of the participants in study \cite{DBLP:conf/icra/Abeysirigoonawardena19}.
Standard BO has been known theoretically to suffer from scalability for high dimensional space, so genetic algorithm (GA) is attracted more attention. 
AV-Fuzzer \cite{DBLP:conf/issre/LiLJTSHKI20} designs a local fuzzer that increases the exploitation of local optima in the areas where highly likely safety hazardous situations are observed.
The study \cite{DBLP:journals/corr/abs-2106-00873} adopts a coverage-driven fuzzing technique to automatically generate diverse configuration parameters to form new driving scenes.
DriveFuzz \cite{DBLP:conf/ccs/KimLRJ0K22} automatically generates and mutates driving scenarios based on diverse factors leveraging a high-fidelity driving simulator.
The study \cite{DBLP:conf/icse/AbdessalemNBS18} uses a learnable evolutionary algorithm to test vision-based control system.
To find more diverse key scenarios, the above methods 
repeat the optimization with significantly different starting points in the search space.

Recently, deep reinforcement learning has been adopted to optimize the control strategy of participants to generate different scenarios.
D2RL \cite{feng2023dense} trains a policy modeled by a neural network to learn when to execute what adversarial maneuver, which aims to improve testing
efficiency and ensure evaluation unbiasedness.
However, the generated scenarios are sensitive to the chosen weight function.

\section{Background}\label{sec:problem}
\noindent
We proceed to introduce the background of the safety-critical search problem, including the problem statement following definitions and the framework overview.

\subsection{Problem Statement}
\noindent
We study the problem of finding the safety-critical scenarios for ADS testing, which consists of a set of variables and a behavior tree that manages all participants' behaviors. 

\begin{definition}[\textbf{Variable}]
	A variable varies with different values for diverse ADS simulations. 
	An $\mathcal{LS}$ may contain multiple variables, and a vector of specific values refers to a $\mathcal{CTS}$.
	Note that, the specific value of a variable is sampled from normal distribution, uniform distribution with range, and user-defined discrete values.
\end{definition}

In fact, there are dependencies between variables. For example, the ego must be faster than the participant in an overtaking scenario.
We define an $relative\quad variable$ to denote the relation between variables $var_j$ and $var_i$, which is formalized:
\begin{equation}\label{eq:rel_var}
	var_j = f(var_i),
\end{equation}
where $f(\cdot)$ is a user-defined function. $var_j$ depends on $var_i$ when $f(\cdot)$ is chosen.
Empirically, an $relative\quad variable$ can reduce the input space by 1 order of magnitude.

\begin{definition}[\textbf{Behavior Node}]
	A behavior node contains a behavior trigger condition and a specific behavior.
	The condition is used to decide when to execute the behavior, while the behavior denotes a specific execution action to constraint the trajectory. 
	Behavior nodes are divided into non-leaf nodes and leaf nodes. The non-leaf nodes are used to describe the execution order, including sequence, parallel, cyclic, and sequential selection. 
	The leaf nodes are executable behaviors.
\end{definition}

We define 6 trigger conditions: \textsf{time}, \textsf{distance}, \textsf{area}, \textsf{relative position}, \textsf{ends by behavior}, and \textsf{combined conditions}.
There are 6 specific behaviors: \textsf{track}, \textsf{changelane}, \textsf{cruise}, \textsf{follow log}, \textsf{merge in}, and \textsf{merge out}.
Note that, the \textsf{cut in} behavior is a specific \textsf{changelane}.

\begin{definition}[\textbf{Behavior Tree ($\mathcal{BT}$)}]
	A behavior tree ($\mathcal{BT}$) are composed of behavior nodes, which drives the participants. 
	We can use different nodes to construct multiple $\mathcal{BT}s$ of participants, which are formed the $\mathcal{BT}$ of the entire traffic scenario.
\end{definition}

Everything is a variable in autonomous driving, only safety is constant. 
The properties of the $\mathcal{BT}$ can be described as variables.
Additionally, the initial state of the participants and ego, the types of participants, weather and so on, can be variables. 
Consequently, it is scalable to a variety of driving environments for diverse ADS testing.

\noindent
\textbf{Problem Definition.} 
Given an $\mathcal{LS}$ described by a $\mathcal{BT}$, indicating that it contains at least one variable, 
our goal is to quickly search all the realistic safety-critical $\mathcal{CTS}s$ where the participants behave normally. 
This is formalized as follows:
\begin{equation}\label{eq:search}
	\begin{aligned}
		&arg\max_{x \in  \mathcal{X}} g(x),\\
		s.t. \quad & The\quad participants\quad behave\quad normally.
	\end{aligned}
\end{equation}
where $\mathcal{X}$ is the input space, $x=[var_1,var_2,...,var_n]$ represents a $\mathcal{CTS}$ with $n$ variables. $g(x)$ is a fitness function to compute a simulated score. In the paper, the higher the score, the more critical the scenario. We detail $g(x)$ in Section \ref{sec:fitness_functions}.


\subsection{Framework Overview}

\begin{figure}[tbp]
	\centering
	\includegraphics[width=.5\textwidth]{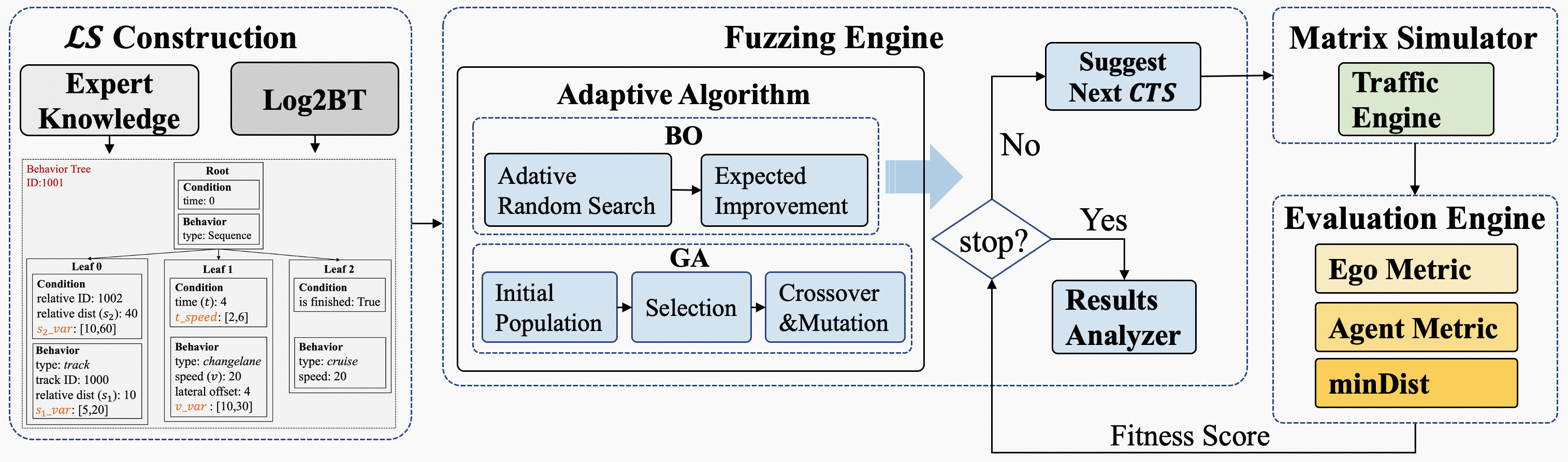}
	\caption{The Matrix-Fuzzer framework.}
	\label{fig:framework}
\end{figure}
\noindent
To solve the above problem, we design an ADS testing framework Matrix-Fuzzer, as shown in Fig. \ref{fig:framework}.
We enhance the basic framework (shown in Fig. \ref{fig:base_framework}) with the following four parts: $\mathcal{LS}$ construction, Fuzzing engine, Matrix simulator, and Evaluation engine.
At first, we have two ways to construct a $\mathcal{BT}$-based $\mathcal{LS}$. The one is based on expert knowledge, while the other adopts the proposed $log2BT$ technology to convert the real-world logged trajectory into a $\mathcal{BT}$.
In each iteration, the Fuzzing engine adopts an optimization algorithm to sample a vector of variable values to generate a $\mathcal{CTS}$. 
Our Matrix simulator executes the input $\mathcal{CTS}$ and the Evaluation engine feeds the fitness score back to the Fuzzing engine for the next iteration.
During this period, the Fuzzing engine records the results of each simulation.
The process is repeated until the stop condition is reached. 
Finally, the result analyzer in Fuzzing engine classifies and analyzes the root causes of the safety-critical scenarios.



\section{Method}\label{sec:method}
\noindent
We proceed to introduce the ADS testing framework Matrix-Fuzzer, including $\mathcal{LS}$ construction, Fuzzing engine, Matrix simulator, and Evaluation engine.

\subsection{$\mathcal{LS}$ Construction}
\noindent
There are two ways to construct a $\mathcal{BT}$-based $\mathcal{LS}$. 
The former is base on expert knowledge, while the latter is $log2BT$.

\subsubsection{Knowledge-based $\mathcal{LS}$ Construction}\label{sec:knowledge}
\begin{figure}[htbp]
	\centering
	\includegraphics[width=0.45\textwidth]{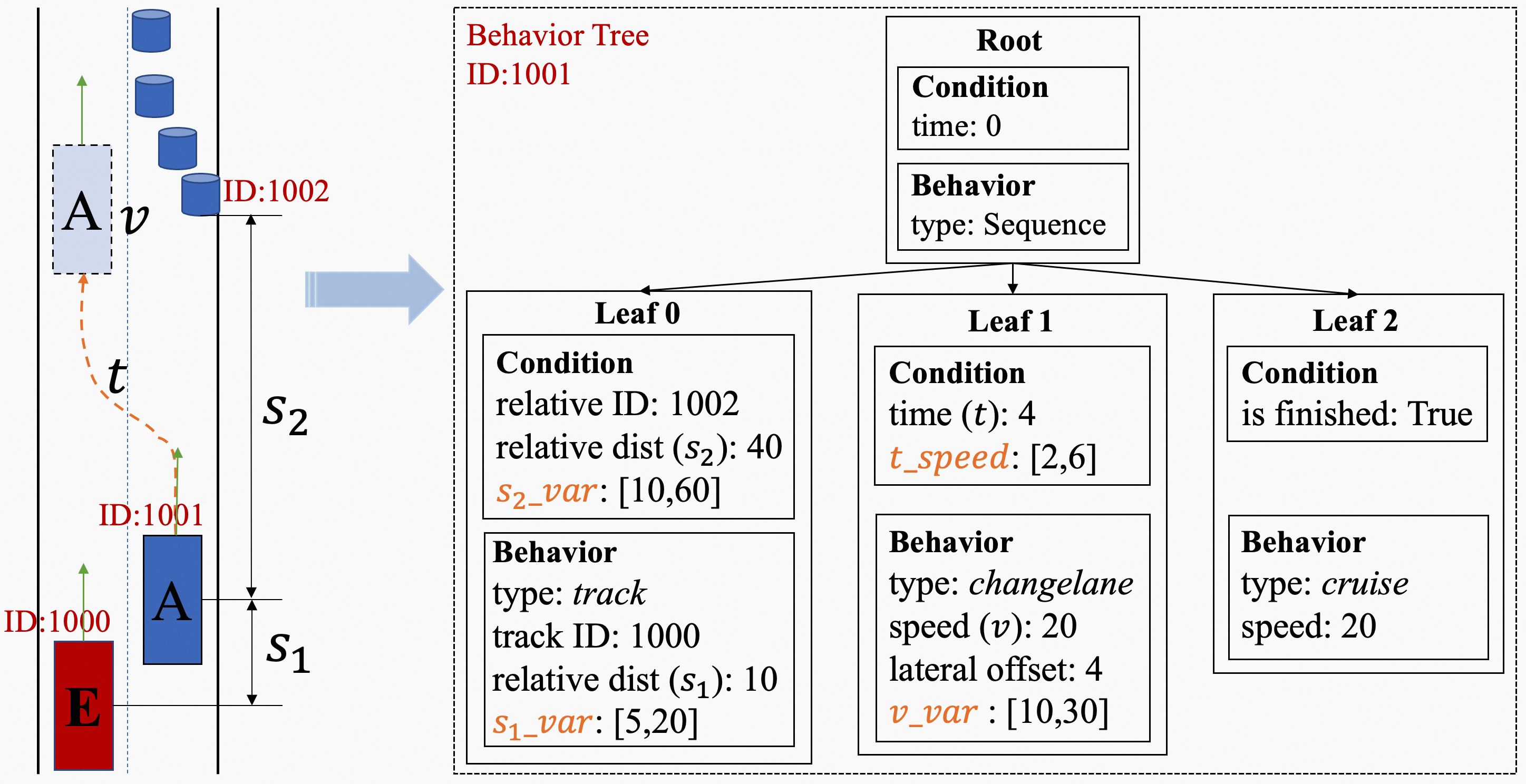}
	\caption{An example of expert knowledge-based $\mathcal{LS}$ construction. (Red vehicle denotes the ego (E) and blue vehicle denotes the agent (A)).}
	\label{fig:cutin}
\end{figure}

Existing approaches demonstrate the significance of domain-specific knowledge in ADS testing.
On the one hand, these approaches rely on expert knowledge to replicate accidents from the scenario database \cite{DBLP:journals/access/RiedmaierPLSD20}, thus validating ADS's ability to handle known risks.
On the other hand, testing the ADS's adaptability to unfamiliar driving environments requires the construction a realistic $\mathcal{LS}$ given an $\mathcal{FS}$ description.
Give an artificial scenario here.
\begin{exmp}[Virtual cut in]\label{exmp:cutin}
	An $\mathcal{FS}$ describes that the nearby agent observes a road construction ahead, then cuts in the ego, and finally cruises.  
	The corresponding $\mathcal{LS}$ expresses that the agent is tracking $s_1$ ahead of the ego, and it suddenly observes a road construction ahead.
	In response, the agent makes a lane change action at a distance of $s_2$ from the construction area, taking $t$ time to reach the end speed of $v$ before cruising.
	This scenario is refined into the three behaviors: \textsf{track}, \textsf{changelane}, and \textsf{cruise}. 
	The corresponding $\mathcal{BT}$-based $\mathcal{LS}$ is shown in Fig. \ref{fig:cutin}, containing four variables: $s_1$, $s_2$, $t$, and $v$, which determine the agent's behavior. 
	Note that, we omit some unimportant attributes here.
\end{exmp}


\begin{figure}
	\centering
	\subfigure[Convert raw trajectory to a $\mathcal{BT}$.]{\includegraphics[width=\linewidth]{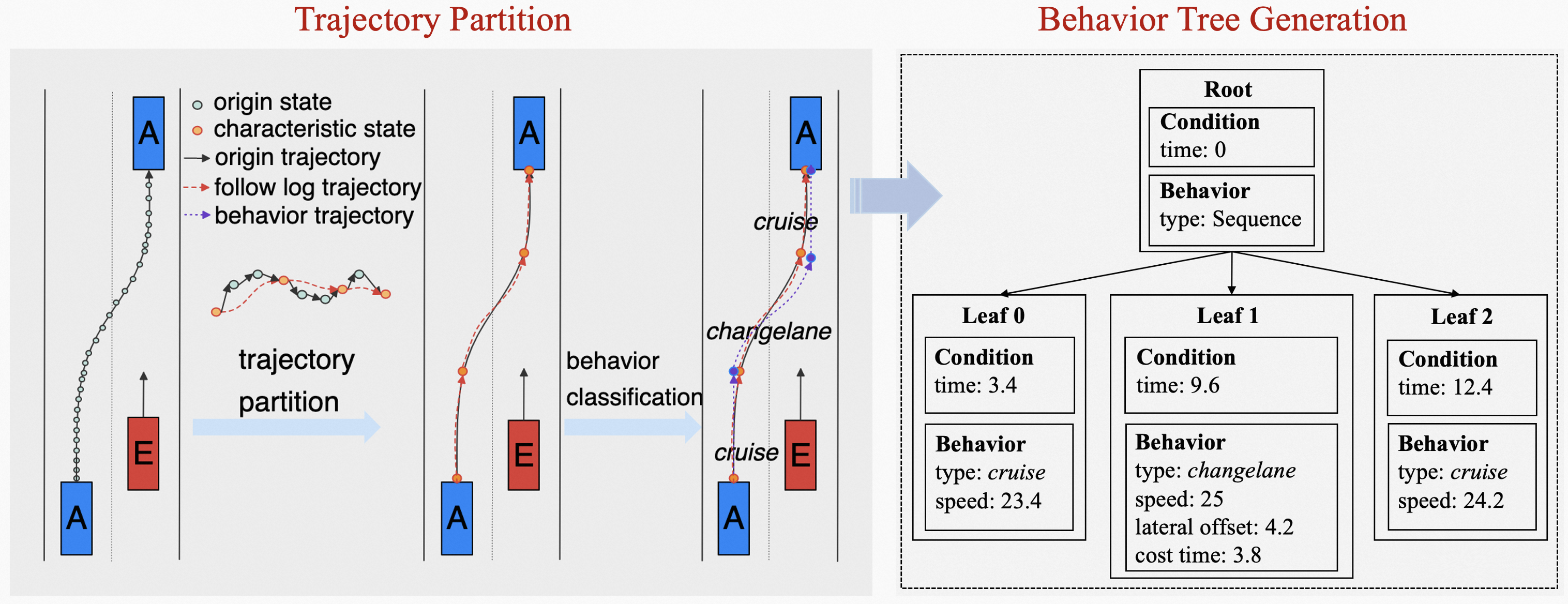}\label{fig:log2bt}}
	\subfigure[Set different variable ranges from statistical distributions.]{\includegraphics[width=\linewidth]{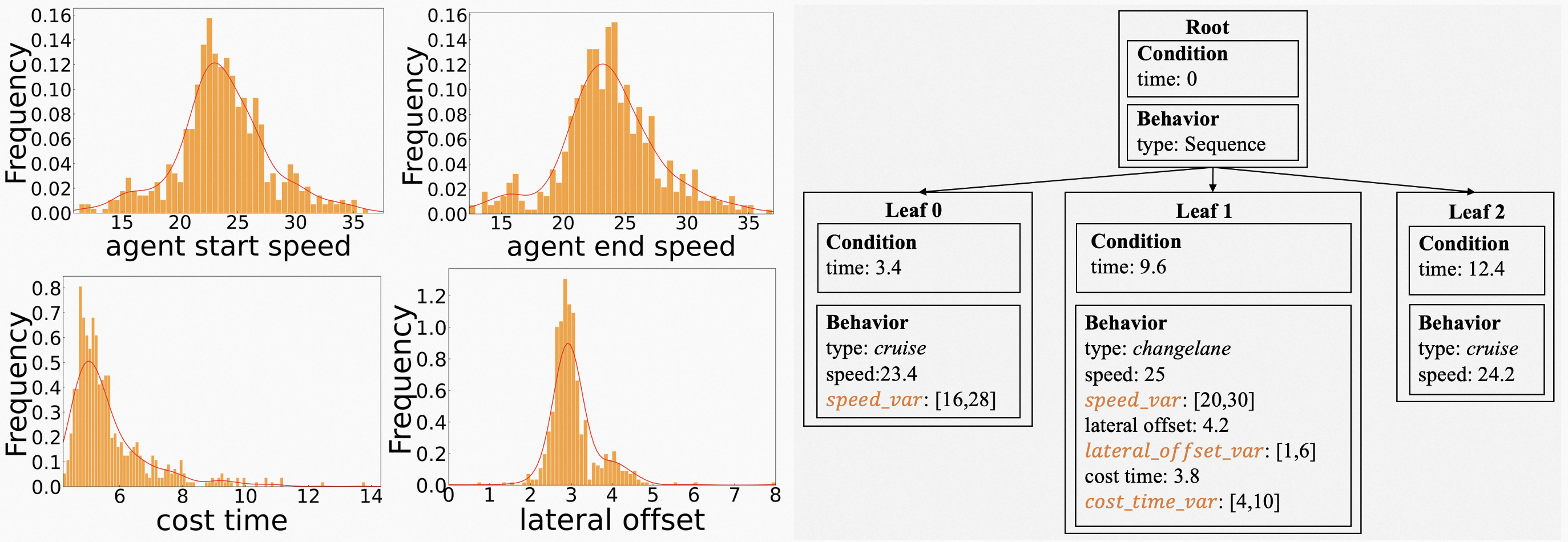}\label{fig:log2bt_LS}}
	\caption{The $log2BT$ workflow.}\label{fig:log2LS}
\end{figure}

\subsubsection{$log2BT$}\label{sec:log2bt}
Inevitably, the knowledge-based $\mathcal{LS}$ deviates from that of NDE under over-reliance on manual expertise.
Choosing an appropriate range for variables is a significant challenge. 
In the above Example \ref{exmp:cutin}, the agent is unable to change lanes in time, and consequently collides with the construction area when the distance $s_2$ is relatively small. 
Although this scenario is invalid, it is often overlooked in existing methods.

To address this problem, we propose $log2BT$ technology to abstract logged road-users' trajectories to behavior sequences. 
Furthermore, we adjust the properties of behavior trees by setting variable ranges with realistic driving distributions. 
Fig. \ref{fig:log2LS} illustrates the $log2BT$ workflow on the following scenario. 
\begin{exmp}[Real cut in]\label{exmp:log_cutin}
	An agent cuts in the ego in the real-world driving environment. 
	Specifically, the agent first cruises at speed $v_1$. Then, it cuts in the ego at time 3.4s. It takes $t$ to cross road with lateral offset $lat$, and finishes the \textsf{change lane} behavior at speed $v_2$.
	Fig. \ref{fig:log2bt} illustrates the process of converting raw log to a $\mathcal{BT}$.
	Fig. \ref{fig:log2bt_LS} illustrates that we set reasonable ranges for different variables under the statistical variable distributions of \textsf{cut in} behavior.
\end{exmp}

Specifically, our $log2BT$ consists of two primary steps: trajectory partition and behavior tree generation. 
A comprehensive outline of the $log2BT$ is detailed in Algorithm \ref{alg:log2bt}.

\noindent
\textbf{1. Trajectory Partition}: To reduce the dimensionality of the original logged trajectories, we cluster and partition the trajectories.
The approaches \cite{DBLP:conf/kdd/GaffneyS99,DBLP:conf/sigmod/LeeHW07} cluster similar trajectories into one group, while
CRISCO \cite{DBLP:conf/kbse/TianWYJ00LY22} partitions trajectory by marking split points with different directions.
However, these methods overlook the velocity and acceleration of points, leading to the omission of significant behavioral characteristics, including acceleration, deceleration, and lane change actions.
To identify potential behaviors, we design a Frenet-based approximate trajectory partition (FATP) algorithm using minimum description length principle.

We first project the trajectory point $p_i = (x_i,y_i,z_i,t_i)$ into the state ${State}_i=(s_i,\dot{s_i},\ddot{s_i},d_i,\dot{d_i},\ddot{d_i},t_i)$ given the reference path $\mathcal{F}$ \cite{DBLP:conf/icra/WerlingZKT10}, where $s_i$ ($\dot{s_i}$,$\ddot{s_i}$) and $d_i$ ($\dot{d_i}$,$\ddot{d_i}$) denote the longitudinal displacement (speed, acceleration) and lateral displacement (speed, acceleration) at $i$-th point, respectively.
Then, a planning trajectory (called \textsf{follow log} trajectory in Fig. \ref{fig:log2bt}) is determined from the start state ${State}_i$ to the end state ${State}_j$.
Specifically, the longitudinal trajectory is calculated by a quartic polynomial, while the lateral trajectory is calculated by a quintic polynomial. 
We define the distance between the origin trajectory $\mathcal{\tau}$ and the planning trajectory $\mathcal{\tau}_{part}$ as the partition cost:
\begin{equation}\label{eq:part_cost}
	cost{part} = \sum_{k=i}^{j}||{State}_k-{State}_{k,part}||^2,
\end{equation}
\textsf{FrenetMDLpar()} in Algorithm \ref{alg:log2bt} denotes to calculate $costpart$.
The state ${State}_j$ is identified as a characteristic state ($\mathcal{CS}$) if $costpart$ is greater than the threshold $\epsilon_{part}$.
Like that, each trajectory is divided into several segments by the $\mathcal{CS}s$ (lines 1-15).

\noindent
\textbf{2. Behavior Tree Generation}: To be more comprehensible and editable, we classify the segments to semantic behavior sequence.
A rule-based behavior pattern identification is applied (lines 16-27).
In particular, for each segment $(\mathcal{CS}_{i},\mathcal{CS}_{i+1})$, it denotes \textsf{change lane} behavior when the lateral offset is greater than the threshold $\epsilon_{lat}$, while it is identified as \textsf{cruise} behavior when the velocity variation is less than the threshold $\epsilon_{vel}$.
Otherwise, it retains \textsf{follow log} behavior to ensure good trajectory reconstruction. 
Furthermore, we build corresponding behavior node and add them to a $\mathcal{BT}$ sequentially. 
Finally, the $\mathcal{BT}$ driving the agent is constructed.
Note that, the properties of behaviors constraint on the agent's trajectory (called behavior trajectory in Fig. \ref{fig:log2bt}) automatically.

Furthermore, the properties of the $\mathcal{BT}$ are described by variables to generalize the real-world scenario.
We mine the driving distributions from NDE and set variables with reasonable ranges shown in Fig. \ref{fig:log2bt_LS}.
We believe that $log2BT$ technique opens a direction for the efficient utilization of raw scenarios, not just for virtual records. 
As the storage overhead is reduced by at least 100x after converting the raw log to a $\mathcal{BT}$, and the potential test value for ADS is huge to construct an $\mathcal{LS}$ via $log2BT$.
That is, we can not only reconstruct the origin scenarios, but also mine the potential critical events from NDE.

\begin{algorithm}[h]
	\caption{$log2BT$}\label{alg:log2bt}
	\KwIn{A trajectory $\mathcal{\tau}$, Frenet Frame $\mathcal{F}$}
	\KwOut{A behavior Tree $\mathcal{BT}$}

	\tcc{\underline{\textbf{1. Trajectory Partition}.}} 
	Initialize a set $\mathcal{CS}s$ of characteristic states.\\
	Project Trajectory $\mathcal{\tau}$ via $\mathcal{F}$ to obtain $\mathcal{S}tates$.\\
	Add ${State}_1$ into the set $\mathcal{CS}s$.\\
	$startIndex := 1, length := 1$\\
	\While{$startIndex + length \leq n$}
	{$currIndex := startIndex + length$\\
		$costpar := FrenetMDLpar(pstartIndex; pcurrIndex)$\\
		\eIf{$costpar > \epsilon_{part} $}{Add $State_{pcurrIndex_{¡}-1}$ into the set $\mathcal{CS}s$\\
		$startIndex := currIndex_{¡-1}, length := 1$
			}
			{$length := length + 1$}
				}
	Add ${State}_n$ into the set $\mathcal{CS}s$.\\

	\tcc{\underline{\textbf{2. Behavior Tree Generation}.}}
	Initialize $\mathcal{BT} \gets \emptyset $ \\
	\For{$i=1$ \KwTo $\mathcal{CS}s.length$}{
		Initialize behavior node $bn \gets \emptyset $ \\
		\uIf{$|\mathcal{CS}_i.d - \mathcal{CS}_{i+1}.d| > \epsilon_{lat}$}
			{$bn \gets$ \textsf{changelane} node with  $(\mathcal{CS}_{i},\mathcal{CS}_{i+1})$}
		\uElseIf{$|\mathcal{CS}_i.\dot{s} - \mathcal{CS}_{i+1}.\dot{s}| < \epsilon_{vel}$}
		{$bn \gets$ \textsf{cruise} node with $(\mathcal{CS}_{i},\mathcal{CS}_{i+1})$}
		\uElse{$bn \gets$ \textsf{follow log} node with $(\mathcal{CS}_{i},\mathcal{CS}_{i+1})$}
		
		$\mathcal{BT}$.add($bn$)
		}	
	\textbf{Return} $\mathcal{BT}$.\\
\end{algorithm}

\begin{figure*}[tbp]
	\centering
	\includegraphics[width=.95\textwidth]{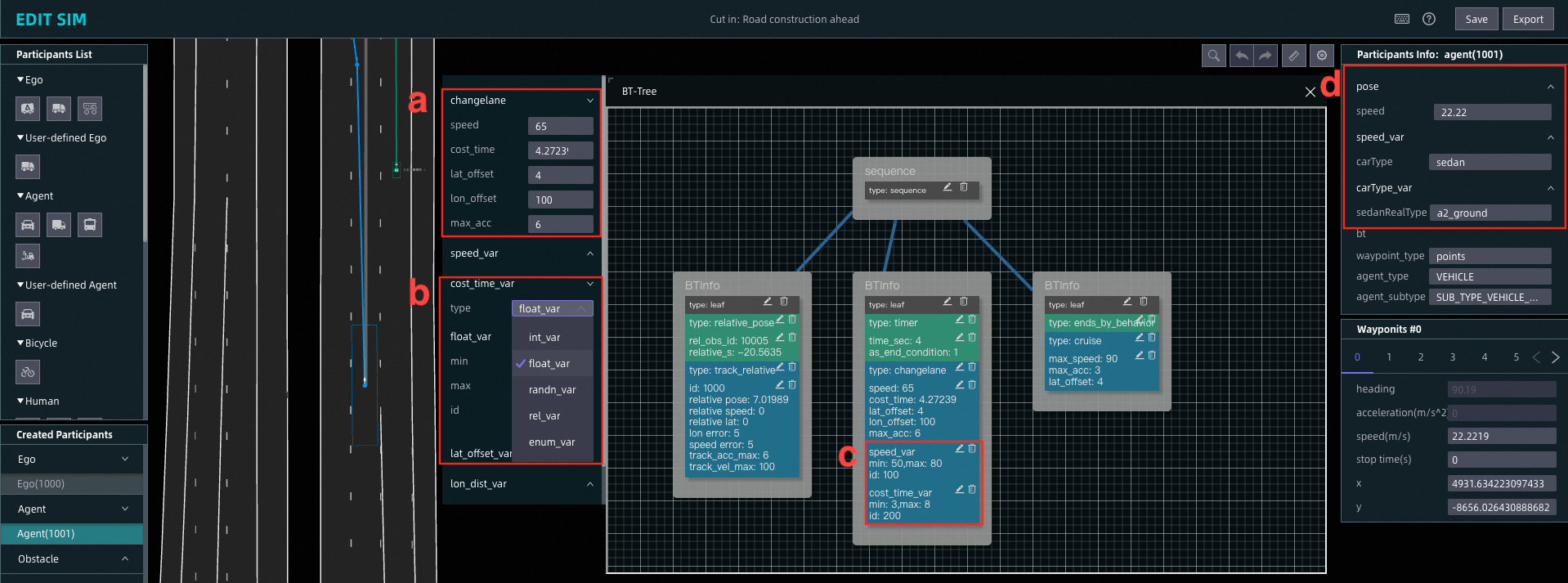}
	\caption{The Matrix Simulator Editor. It allows us to define diverse driving environments with different intersections. 
	\textbf{a} shows the properties of \textsf{change lane} behavior. 
	\textbf{b} denotes the multiple types of variable.
	\textbf{c} shows the property variable with reasonable range.
	\textbf{d} shows the initial state (position, speed, and type).
	}
	\label{fig:matrix}
\end{figure*}

\subsection{Fuzzing Engine}
\noindent
BO and GA are popular optimization algorithms to solve the problem formalized in Equation \eqref{eq:search}.
Standard BO has been known to suffer from scalability for high-dimensional space, while the efficiency of GA is affected by the pre-defined number of populations and inheritance.
Surprisingly, we investigate that there is no previous work to compare the efficiency of BO and GA in ADS testing. 
Experimentally, we have insights that BO performs better than GA at low-dimensional space. 
Therefore, Matrix-Fuzzer uses an adaptive optimization algorithm to explore to the input space.
Specifically, we adopt the BO algorithm when the number of variables $n$ is less than the threshold $\epsilon_{n}$ (defaults to 10),
otherwise the GA is applied. Next, we briefly introduce the BO and GA algorithms:

\noindent\textbf{Bayesian Optimization (BO)}: We choose Expected Improvement as the acquisition
function, which accounts for the size of the improvement while exploring and exploiting the function to find a global
maximum.
To explore the input space, we first employ the adaptive random search algorithm \cite{DBLP:conf/icse/AbdessalemNBS18} to conduct random search, and the number of searches is $n*20$.
Adaptive random search is an extension of naive random search that tries to maximize the Euclidean distance between selected points in the input space. 
To avoid falling into a local maximum, the trade-off value of exploration and exploitation is set to 5 to encourage exploration.


\noindent\textbf{Genetic Algorithm (GA)}: 
First, we employ the adaptive random search algorithm \cite{DBLP:conf/icse/AbdessalemNBS18} to randomly initialize a population of size $n*10$ from the input space.
Then, the evaluation engine outputs the fitness scores after the population executes the simulation. They are sorted by the score for next \textsf{Selection/Crossover/Mutation} operations.
The goal of the \textsf{Selection} process is to eliminate unqualified candidates from the population. 
We adopt the roulette to select the winner according to the proportion of their fitness scores. 
If a scenario has a higher score, it will be selected more often.
The selected parents are used to the next \textsf{Crossover} step. 
That means, we randomly swap the values of the variables in the two parent scenarios to get a new test scenario.
To promote the diversity of later generation, we set the mutation rate to 0.5.

Note that, BO and GA both iteratively sample from the input space, 
and select the most suitable parameters of variables according to the fitness score. 
These variables are ultimately expected to move towards the critical region. 

\subsection{Matrix Simulator}
\noindent
To simulate ADS, we use Matrix simulator designed by ourselves.
Matrix simulator provides an interactive scenario editor shown in Fig. \ref{fig:matrix}, enabling user to design diverse driving environments combining with the HD maps.
It allows us to place multiple participants and then execute the scenarios capturing different intersections via the internal Traffic Engine.
There are diverse ways to drive the participants.
Generally, we can define the path of the participants by dragging like Apollo \cite{apollo}.
In addition, editing behavior trees for the participant is convenient.
The $\mathcal{BT}$-based scenario from $log2BT$ (Section \ref{sec:log2bt}) is effortlessly imported for its scalability in ADS testing.
We can vary the types of participants, weather conditions and infrastructures in the test scenarios.

\begin{algorithm}[]
	\caption{Scenario Fitness Score}\label{alg:fit_score}
	\KwIn{A Concrete Test Scenario $\mathcal{CTS}$}
	\KwOut{A score of $\mathcal{CTS}$}

	Choose metric actions for agent and ego.\\
	Evaluate the metrics when $\mathcal{CTS}$ is simulated.\\
	Compute $Score_{agent}$ and $Score_{ego}$  according to Eq. \ref{eq:p_score}.\\
	$Score:=0$\\
	\eIf{Agent behaves unreasonably}
		{$Score:=-Score_{agent}$}
		{\eIf{$Collision() \quad and \quad isResposibility()$}{$Score:=Score_{ego}$}{Compute $Score$  according to Eq. \eqref{eq:fitness_score}.}}
	
	\textbf{Return} $Score$.\\
\end{algorithm}
\subsection{Evaluation Engine}\label{sec:fitness_functions}
\noindent
Criteria aim to guide the search process to find key scenarios.
A scenario is considered critical if a safety measure $g(x)$ (Equation \eqref{eq:search}) is over a user-specified threshold. 
$minTTC$ and $minDist$ are two common measures.
The former represents minimum Time To Collision between
the ego and any of other participants during the simulation. The latter is 
the minimum distance between the ego and any other participants during the simulation.
Although several more fine-grained measures are proposed in \cite{DBLP:conf/ivs/GhodsiHFTTKGA21}, they mainly focus on the reasonability of the ego without considering the participants, guiding the search process to prefer the reckless behaviors of participants.

To reduce generating invalid scenarios, we design adequate criterions to compute the fitness score of the test scenario, which is calculated as follows:
\begin{equation}\label{eq:dist_score}
	Score_{dist} = a*{minDist} + b,
\end{equation}
\begin{equation}\label{eq:fitness_score}
	Score = \alpha_1*Score_{ego} + \alpha_2*Score_{agent} + \alpha_3*Score_{dist},
\end{equation}
where $a$ ($a<0$) and $b$ are the coefficient and bias of $minDist$. $Score_{ego}$, $Score_{agent}$, and $Score_{dist}$ respectively represent the score calculated from the ego, the agent, and the distance.
$\alpha_i(i=1,2,3)$ indicates the corresponding weight of the score item ($\alpha_1>0$, $\alpha_2<0$, $\alpha_3>0$).

To evaluate the rationality of the ego and participants from multiple perspectives, we define $k$ metrics, including collisions, long-term line pressure, aggressive drving, off road and so on.
For $i$-th metric, its score depends on its simulation state, including $success$, $warnning$, and $fail$:
\begin{equation}\label{eq:score_i}
	score_i = \left\{
			\begin{array}{lr}
			0, & success \\
			2, & warning,\\
			5, & fail\\
			\end{array}
	\right.
\end{equation}
Therefore, the score of ego (agent) is calculated:
\begin{equation}\label{eq:p_score}
	Score_{p} = \sum_{i=1}^{k}score_i,p\in\{ego,agent\},
\end{equation}
The fitness score of a $\mathcal{CTS}$ is detailed in Algorithm \ref{alg:fit_score}.
When the agent behaves unreasonably, the fitness score will be negative. 
To reduce invalid collisions, we consider the responsibility for collisions and the scores are high when the collisions are caused by the ego ($isResposibility()$ in Algorithm \ref{alg:fit_score}).
We obtain a positive score when the behaviors of the ego are failed. That denotes a critical safety.
Note that, the metrics for the ego and agent can be different, which depending on the testing requirements.
Therefore, Equation \eqref{eq:fitness_score} is the general formalization of the fitness function expressed in the existing approaches.

\section{Experiments}\label{sec:experiments}

\begin{figure*}
	\centering
	\subfigure{\includegraphics[width=.35\linewidth]{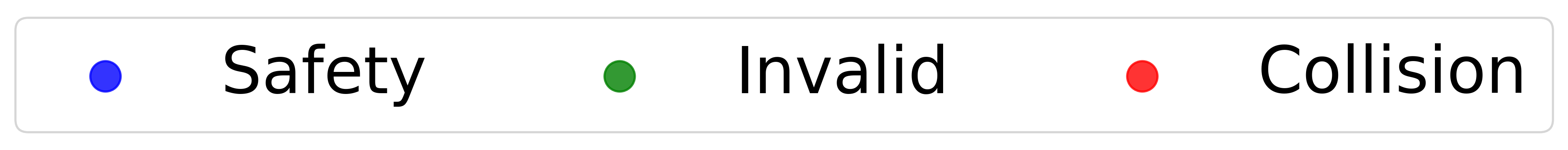}}
	\vspace{-0.2cm}
	\setcounter{subfigure}{0}

	\subfigure[Grid search]{\includegraphics[width=.28\linewidth]{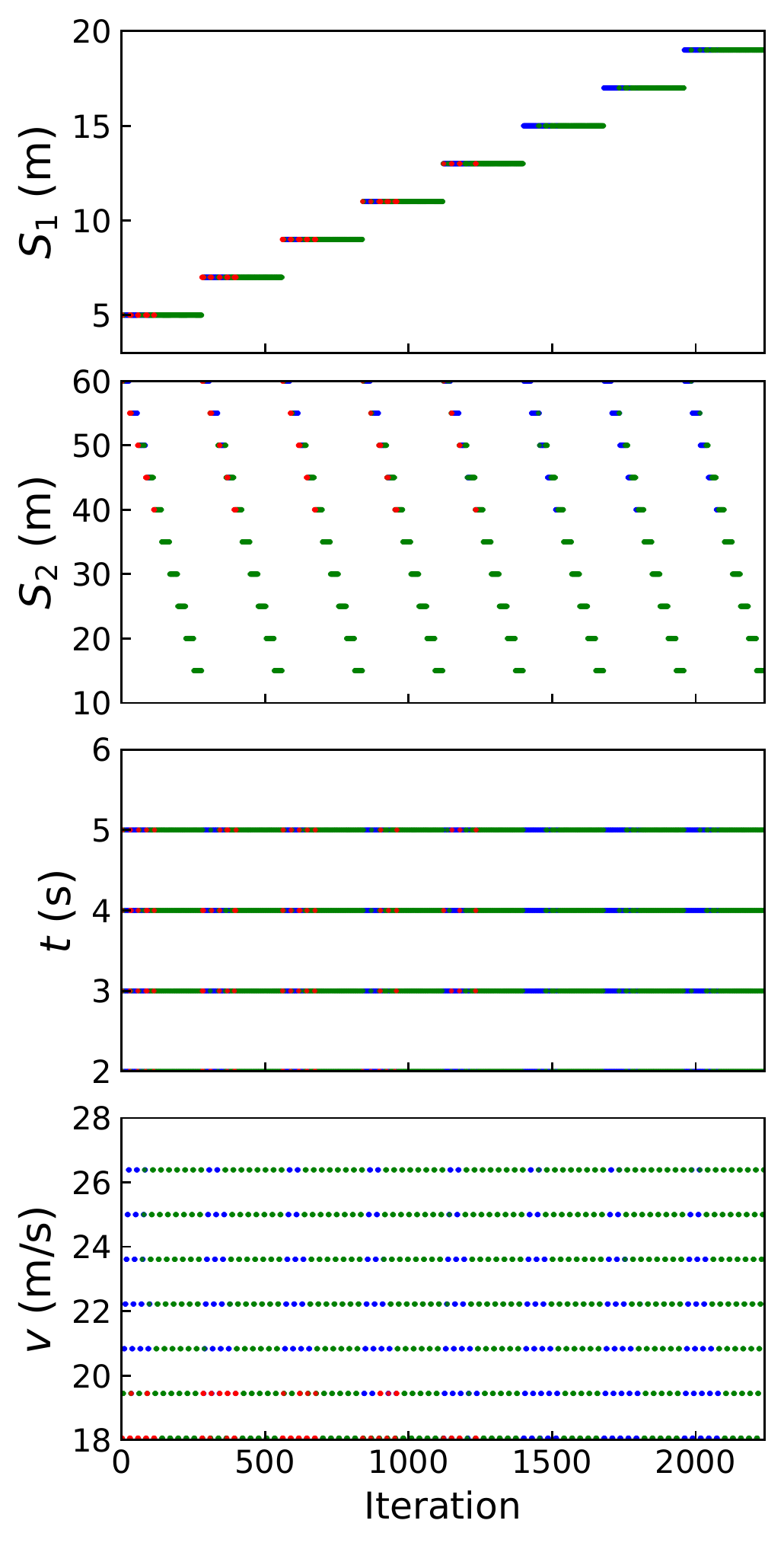}\label{fig:grid_1}}
	\subfigure[AV-Fuzzer]{\includegraphics[width=.28\linewidth]{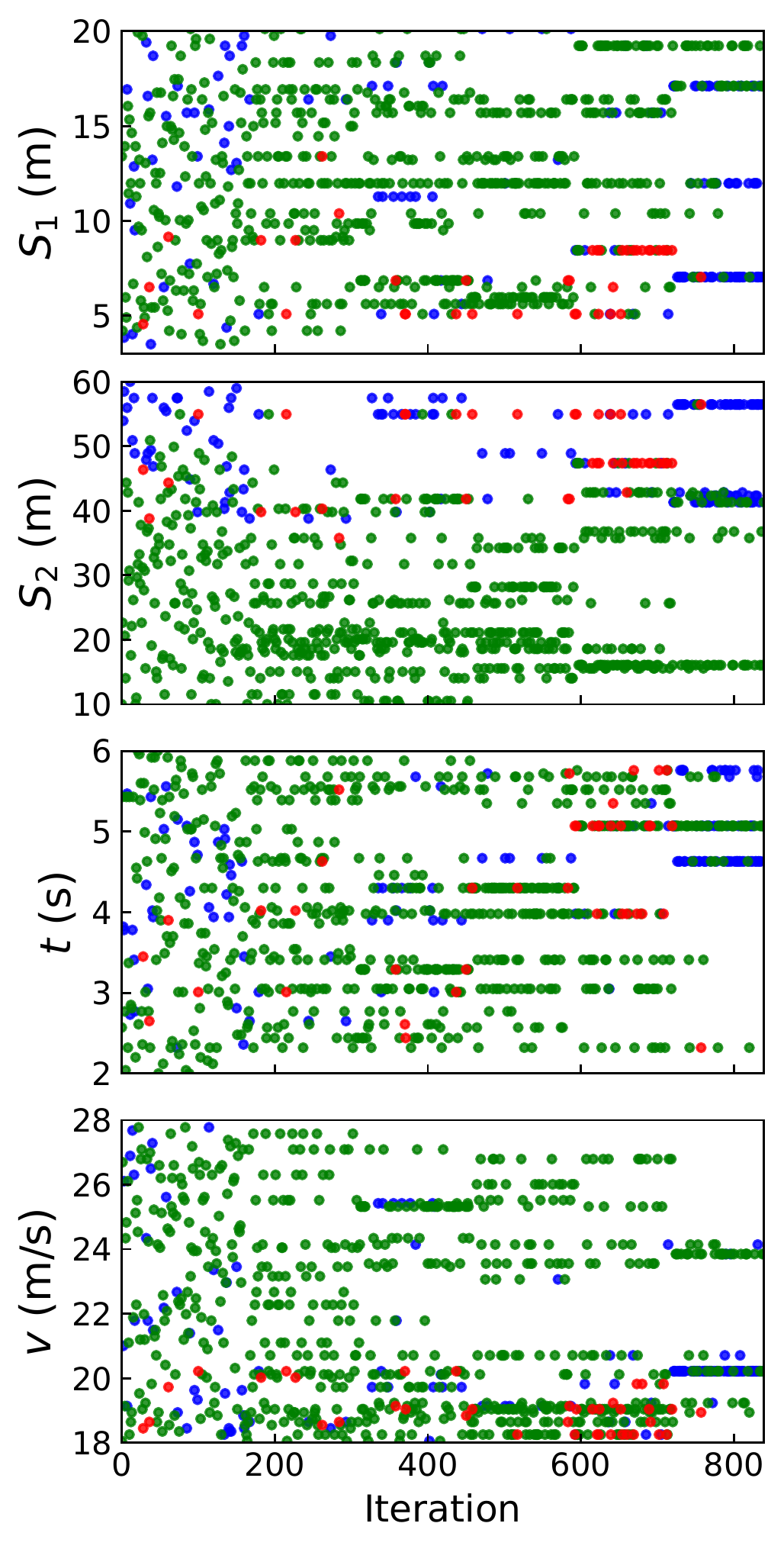}\label{fig:av_fuzzer_1}}
	\subfigure[Matrix-Fuzzer]{\includegraphics[width=.28\linewidth]{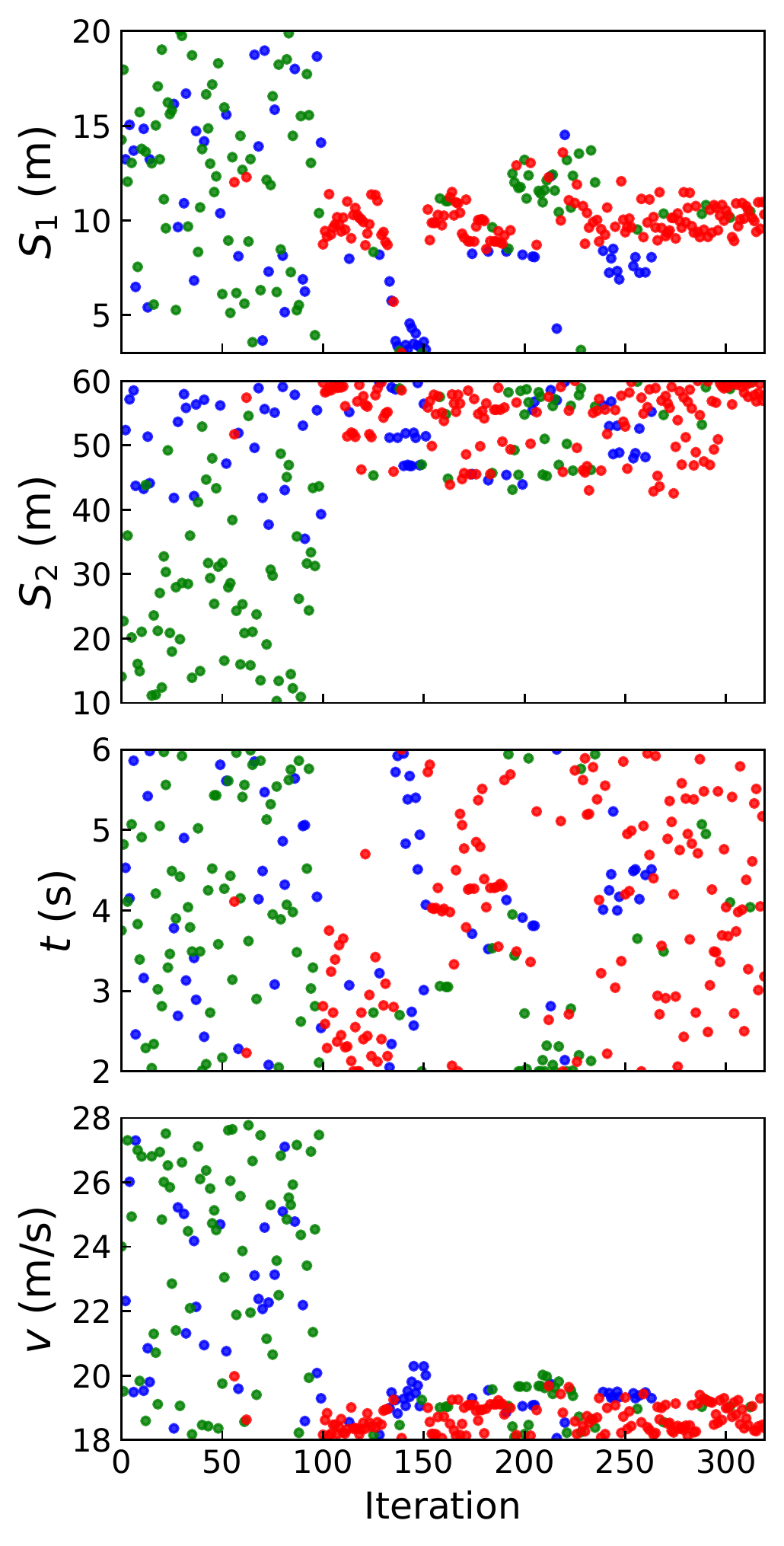}\label{fig:matrix_fuzzer_1}}
	\caption{The search process in Example \ref{exmp:cutin}.}\label{fig:search_process_1}
\end{figure*}

\begin{figure*}
	\centering
	\subfigure[Grid search]{\includegraphics[width=.28\linewidth]{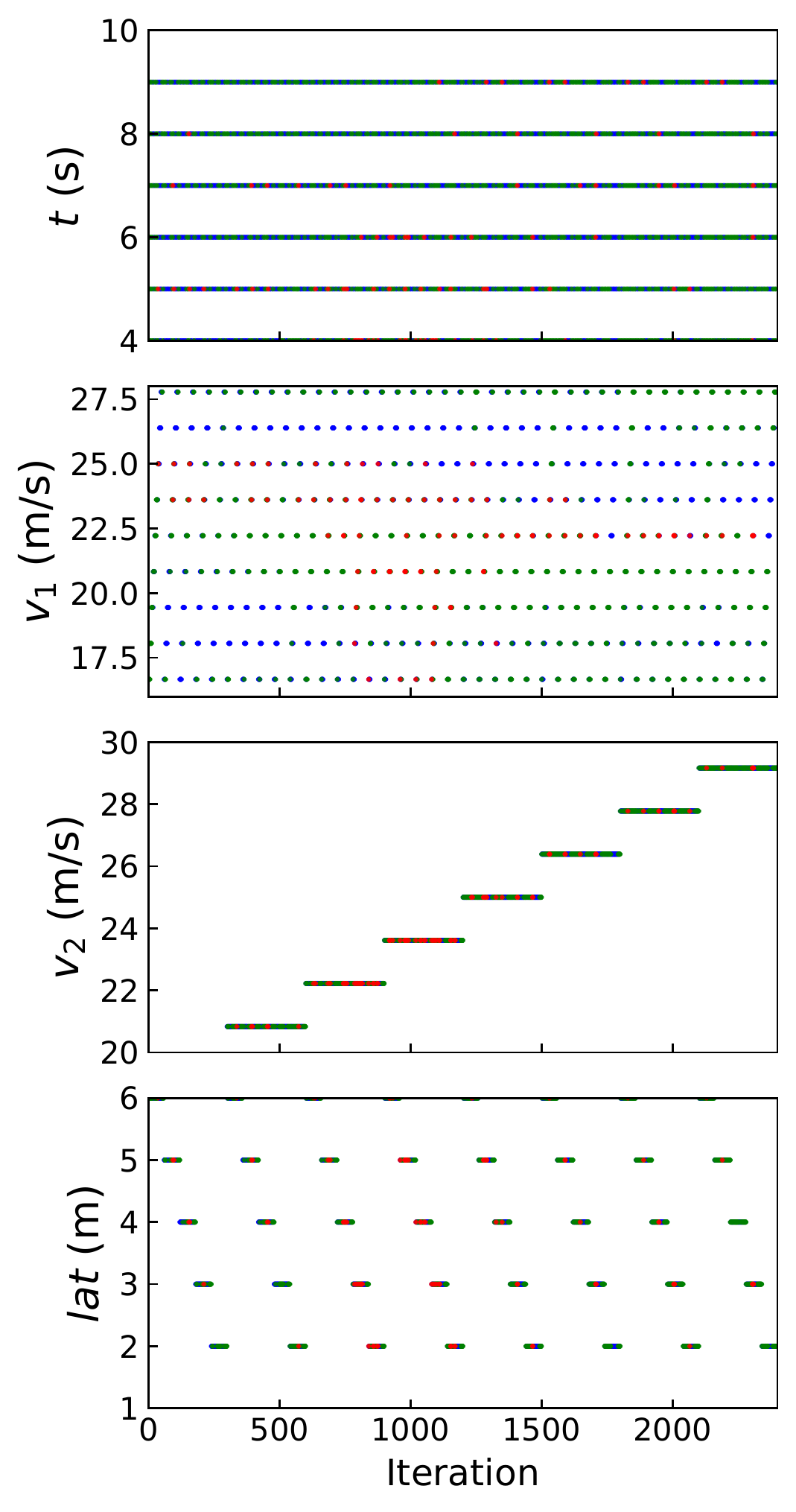}\label{fig:grid_2}}
	\subfigure[AV-Fuzzer]{\includegraphics[width=.28\linewidth]{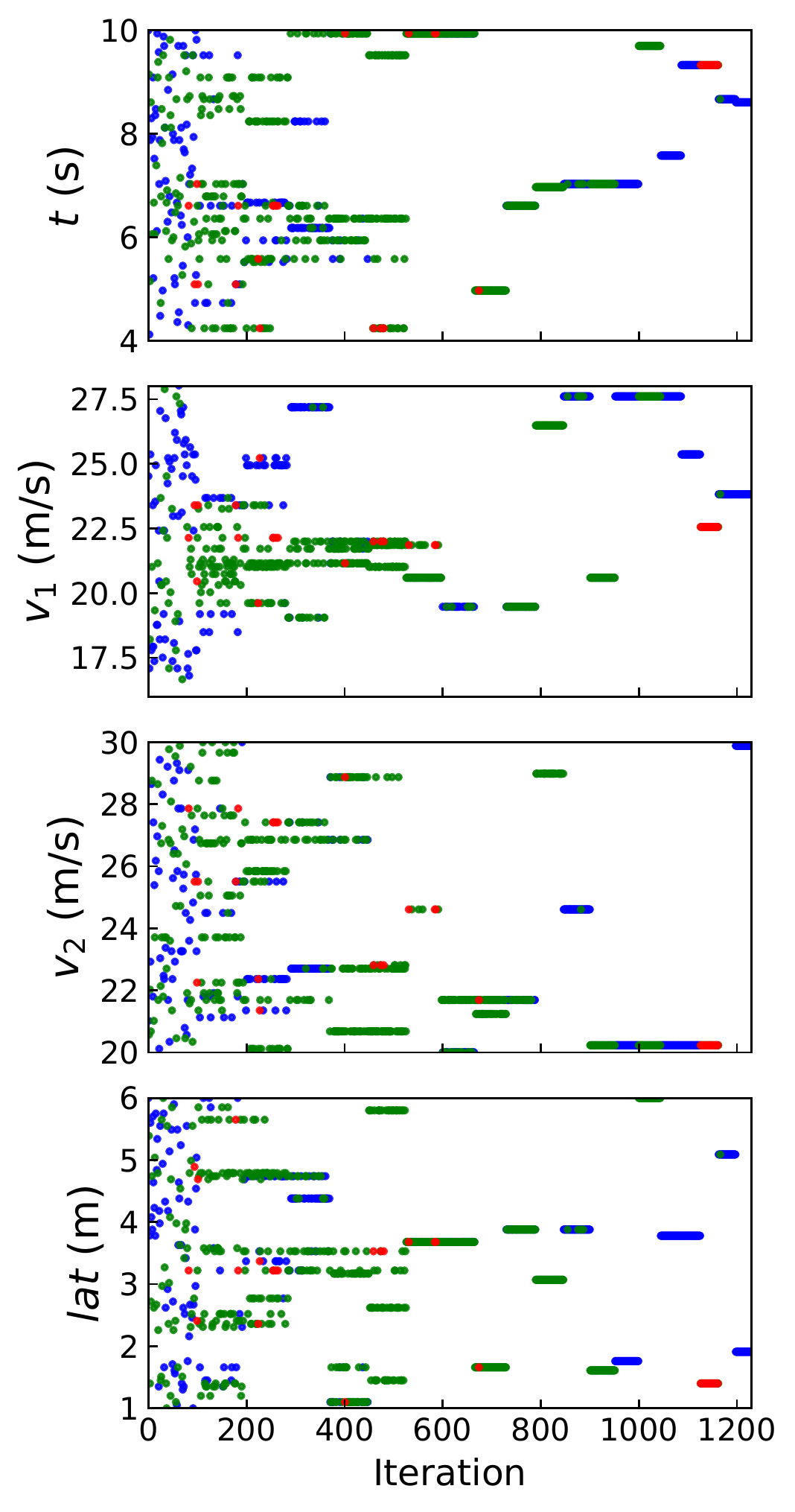}\label{fig:av_fuzzer_2}}
	\subfigure[Matrix-Fuzzer]{\includegraphics[width=.28\linewidth]{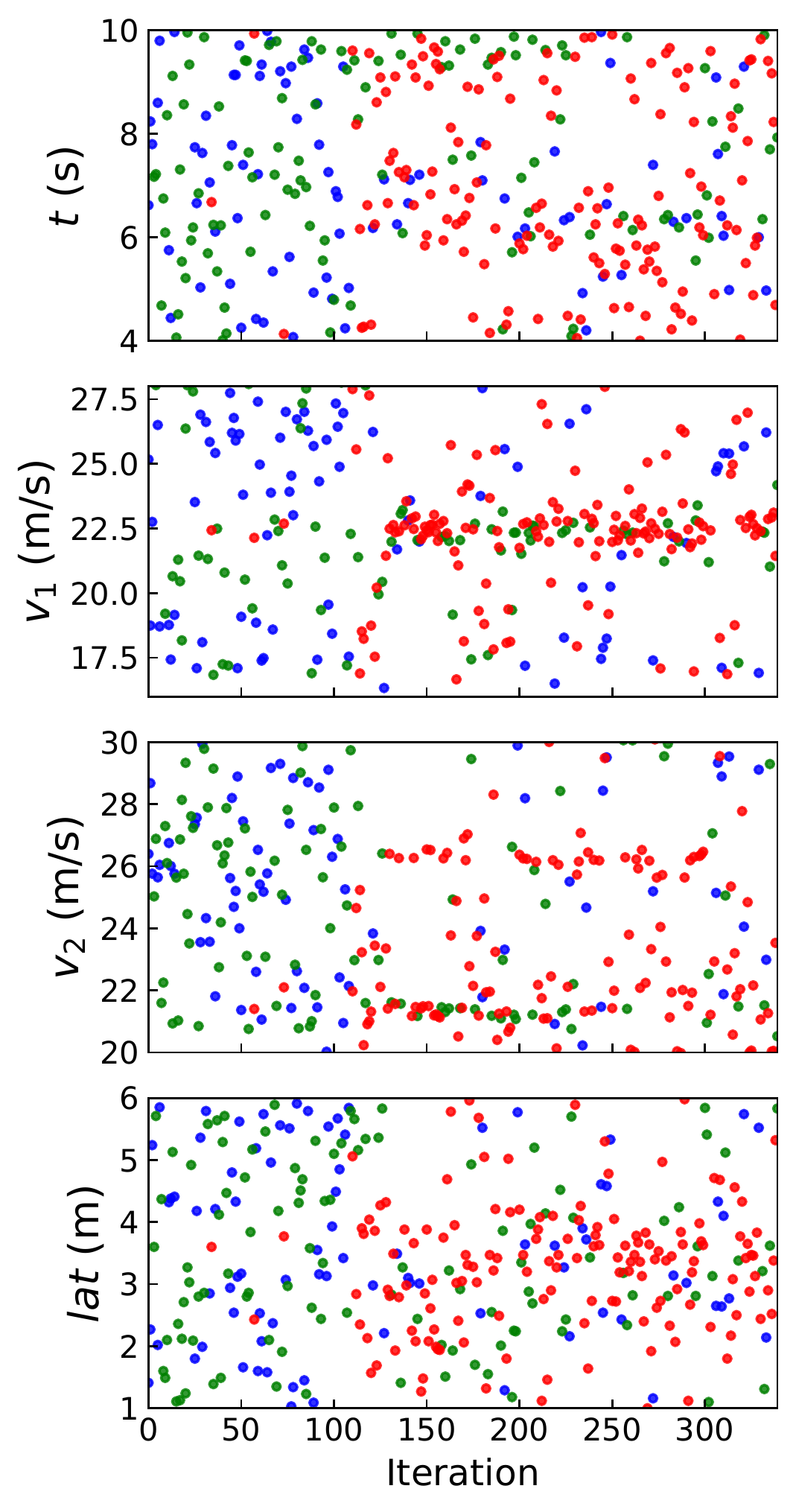}\label{fig:matrix_fuzzer_2}}
	\caption{The search process in Example \ref{exmp:log_cutin}.}\label{fig:search_process_2}
\end{figure*}

\begin{figure}
	\centering
	\subfigure[The search space in 3D]{\includegraphics[width=.75\linewidth]{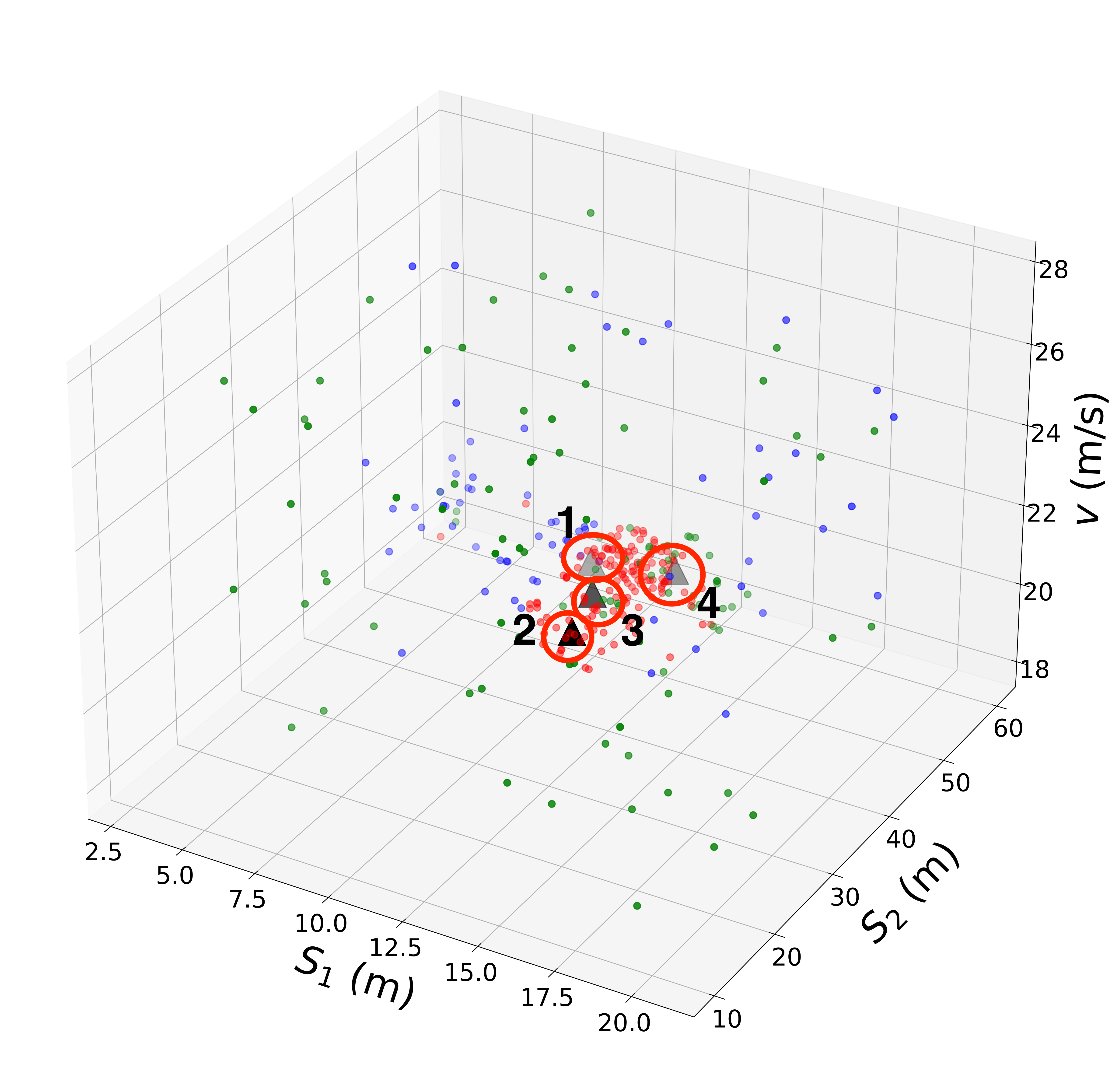}\label{exp:3d}}
	\subfigure[The types of scenarios]{\includegraphics[width=.75\linewidth]{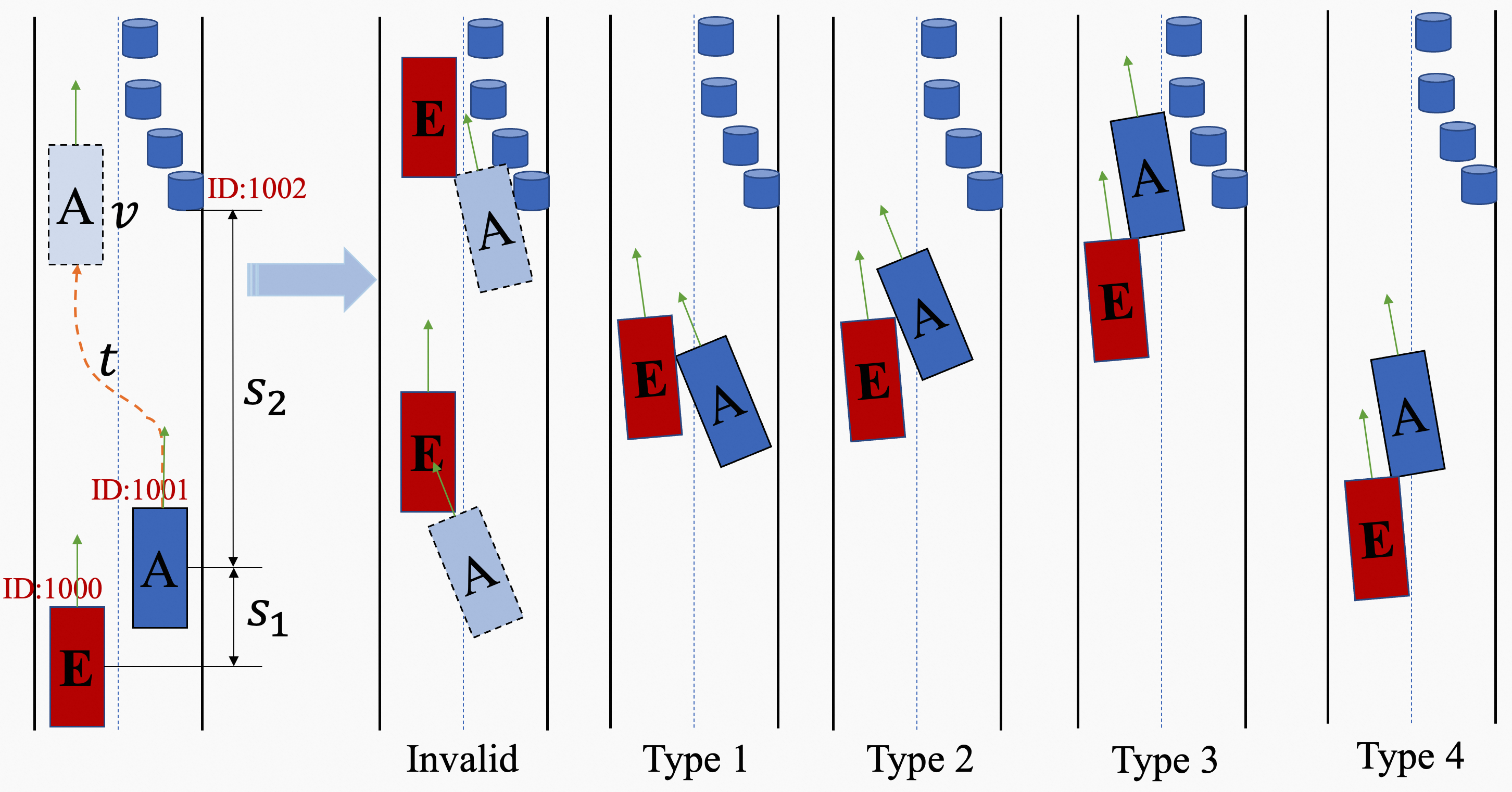}\label{exp:cutin_type}}
	\caption{The types of safety-critical scenarios in Example \ref{exmp:cutin}.}\label{exp:cutin_exp_type}
\end{figure}

\begin{figure*}
	\begin{minipage}[htbp]{0.58\linewidth}
		\centering
		\subfigure[The search process]{
			\includegraphics[width=\textwidth]{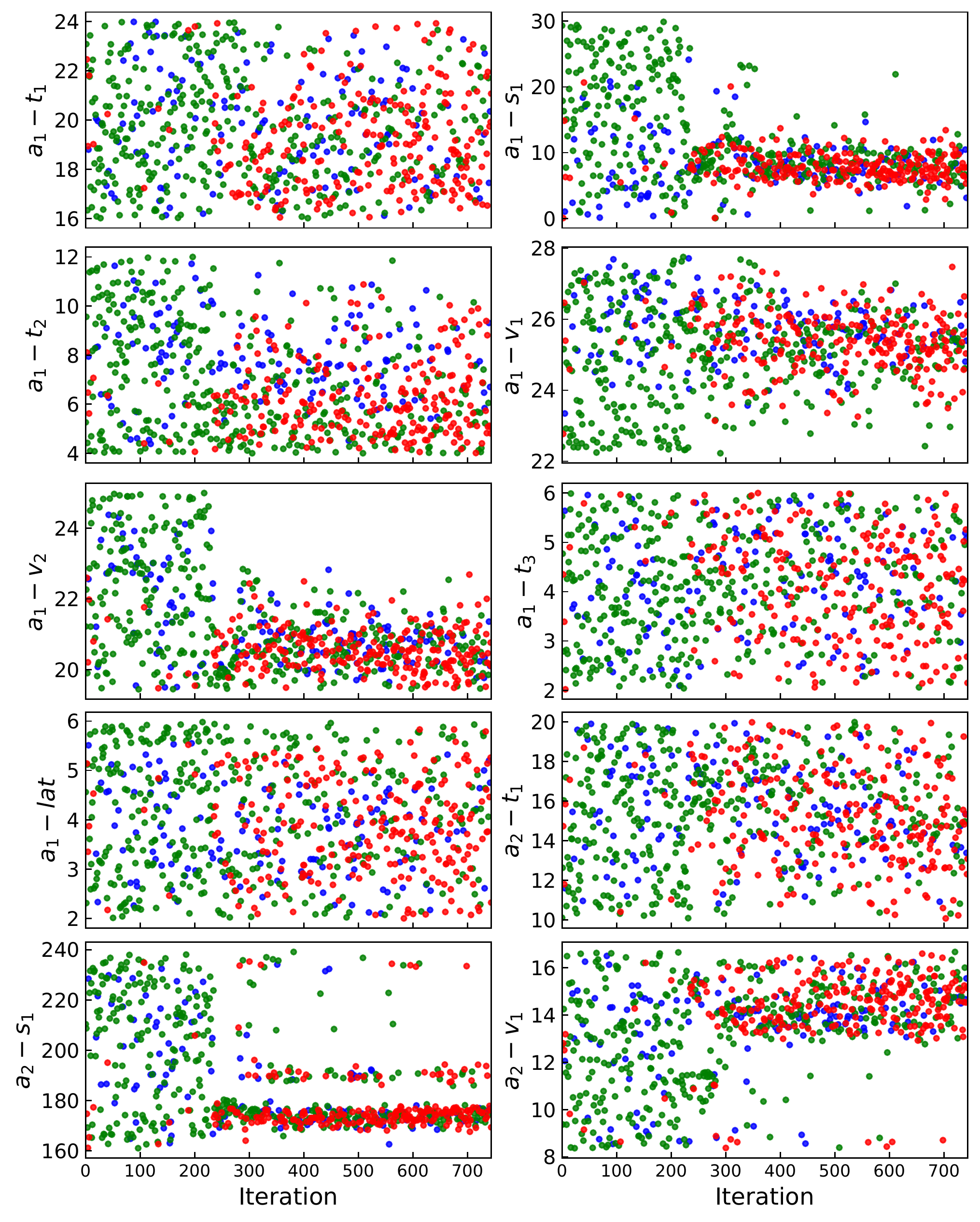}\label{exp:hd}
		}
	\end{minipage}
	\hfill
	\begin{minipage}[htbp]{0.4\linewidth}
		\centering
		\subfigure[The search space in 3D]{
			\includegraphics[width=0.94\textwidth]{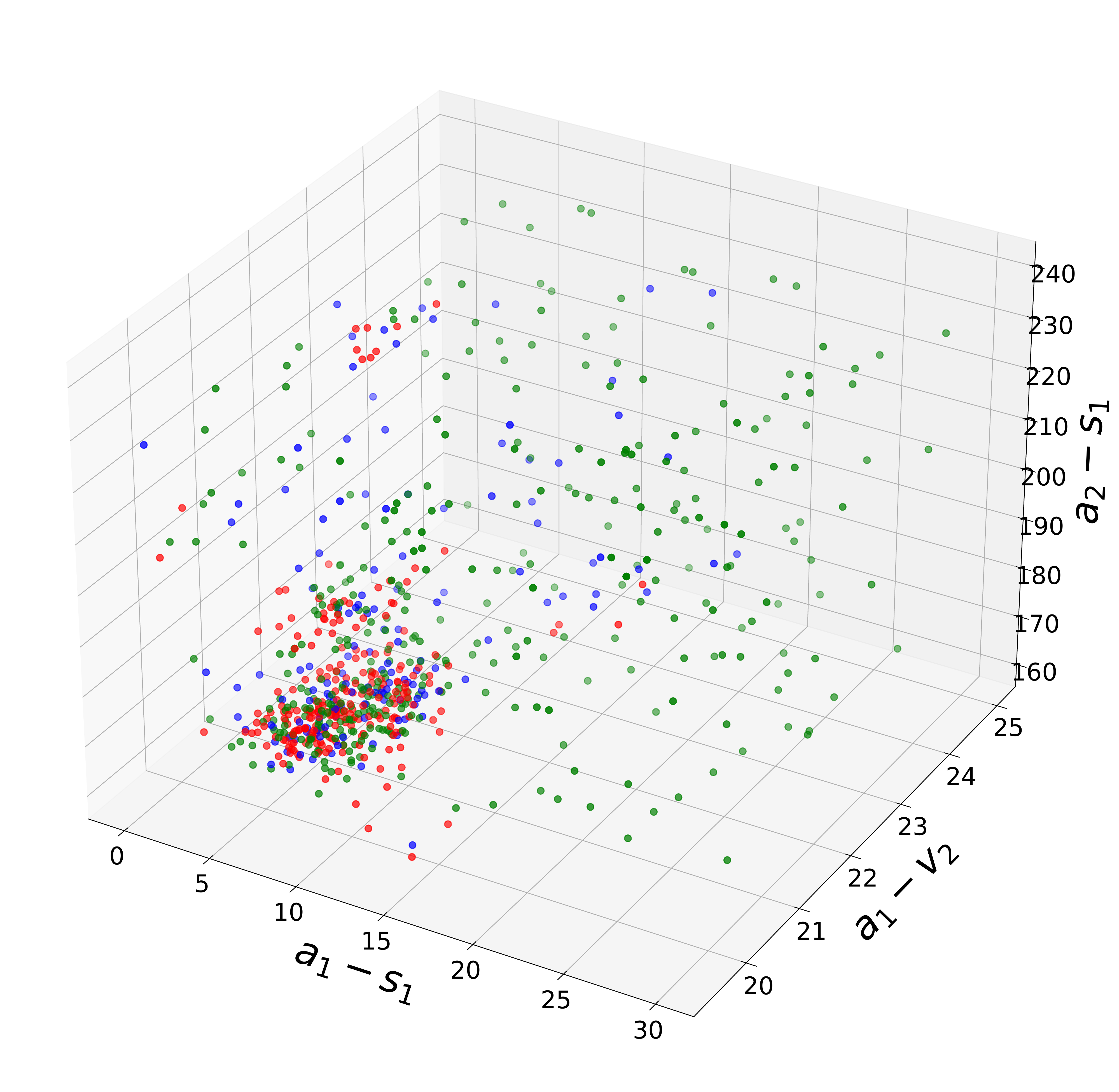}\label{exp:hd_3d}}
		\subfigure[Variable Correlations]{
			\includegraphics[width=0.94\textwidth]{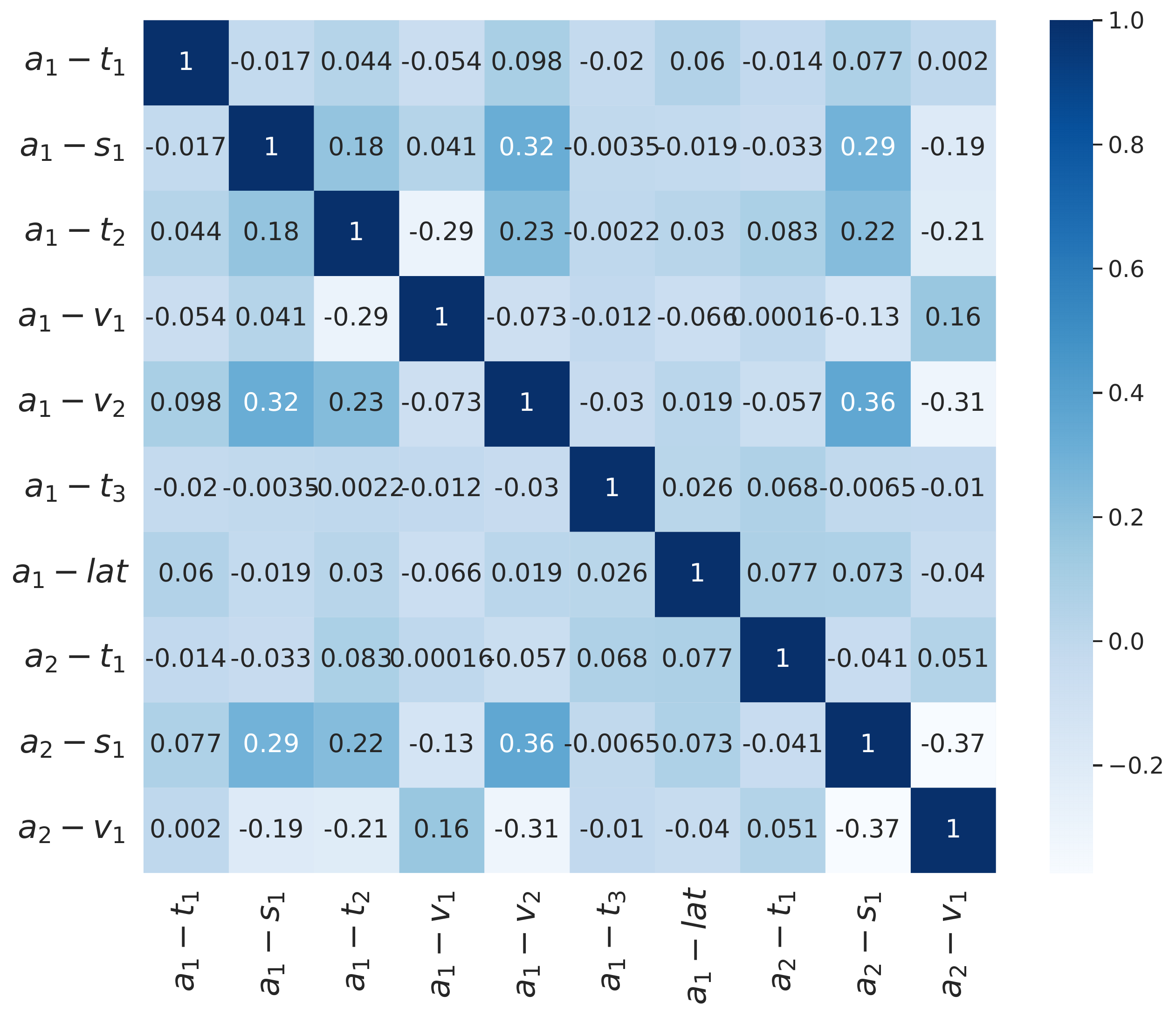}\label{exp:hd_correlation}
		}
	\end{minipage}\label{fig:hd}
	\caption{High-dimensional input space. $a_i (i=1,2)$ denotes the $i$-th agent.}
	\label{fig:hd_search}
\end{figure*}

\noindent
In this section, we apply the Matrix-Fuzzer to generate safety-critical scenarios for testing ADSs on our Matrix simulator.
To evaluate the efficiency and effectiveness of our approach, we answer the following research questions.

\textbf{RQ1:} Is $log2BT$ able to reconstruct the origin trajectories?

\textbf{RQ2:} How effective and efficient is Matrix-Fuzzer to search safety violations from the knowledge-based $\mathcal{LS}$ and the NDE-based $\mathcal{LS}$?
The following aspects need to be compared:
\begin{itemize}
	\item How many safety-critical scenarios are generated?
	\item How many invalid safety-critical scenarios are searched?
	\item How many types of safety violations can be found? 
	\item How distinct are different types of safety violations?
	\item Which variable is critical to the similar safety violations?
\end{itemize}

\textbf{RQ3:} How does Matrix-Fuzzer search safety-critical scenarios in high-dimensional space?
\begin{itemize}
	\item How to choose the search algorithm?
	\item Which variables are critical to the safety violations?
	\item Are there correlations among variables?
\end{itemize}
\subsection{Experiment Design}
\noindent
To answer \textbf{RQ1}, we compare the origin trajectory and the trajectory generated by $log2BT$.
We adopt Average Displacement Error (ADE) that denotes the average
L2 distance between the ground truth state and generated state over all time steps.
In particular, we use $ADE_s$ and $ADE_l$ to measure the longitudinal and lateral errors.
We use two real datasets, called NGSIM \footnote{https://www.opendatanetwork.com/dataset/data.transportation.gov/8ect-6jqj} and our dataset, to evaluate $log2BT$.
As mentioned in Section \ref{sec:log2bt}, we classify the behavior for each characteristic state tuple.
The intuition is to abstract a trajectory segment to a semantic behavior. 
We use $log2BT_{log}$ to denote only \textsf{follow log} behaviors in $\mathcal{BT}s$, while
$log2BT_{semantic}$ distinguishes it with semantic behaviors such as \textsf{changelane} behavior and \textsf{cruise} behavior.

\begin{table}
	\centering 
	\caption{Test $\mathcal{LS}$ Information}
	\label{tb:exp_case_info}
	\resizebox{0.98\linewidth}{!}{
		\begin{tabular}{c|c|c|c|c}
			\hline
			\hline
			Example \ref{exmp:cutin}: $v_{ego}^{init}=22$ & $S_1$:[3,20] & $S_2$:[10,60] & $v$:[18,30] & $t$:[2,6] \\
			Example \ref{exmp:log_cutin}: $v_{ego}^{init}=17$ & $v_1$:[16,28] & $lat$:[1,6] & $v_2$:[20,30] & $t$:[4,10] \\
			
			\hline
			\hline
		\end{tabular}
	}
\end{table}
To answer \textbf{RQ2}, we evaluate Matrix-Fuzzer in comparison with Grid search and AV-Fuzzer \cite{DBLP:conf/issre/LiLJTSHKI20}.
Grid search is an exhaustive search of the input space with the predefined steps, its performance is sensitive to the steps.
AV-Fuzzer applies GA to search for critical scenarios, and repeats the optimization with different starting points in the input space.
Without loss of generality, we demonstrate the experiments on the $\mathcal{LS}s$ described in Example \ref{exmp:cutin} and \ref{exmp:log_cutin}.
Table \ref{tb:exp_case_info} summaries the initial speed of ego and the input spaces.
We default to apply BO in Matrix-Fuzzer.
To better compare with competitors from the aspects mentioned in \textbf{RQ2}, we define the following metrics:
\begin{itemize}
	\item $Total$: the number of total concrete test scenarios.
	\item $Critical$: the number of safety-critical scenarios.
	\item $Invalid$: the number of invalid safety-critical scenarios.
	\item $Types$: the number of types of safety-critical scenarios.
	\item CR: the ratio of $Critical$ to $Total$.
	\item IR: the ratio of $Invalid$ to $Total$.
	\item TR: the ratio of $Types$ to $Total$.
\end{itemize}

Note that, the scenarios are treated as invalid, where the behaviors of participants are unreasonable such as rear end collision with the ego, driving off road, and having collisions with the other participants.

To answer \textbf{RQ3}, we construct the $\mathcal{LS}$ with different number of variables $n$. 
As $n$ grows, the input space grows exponentially, and it's much difficult to search the safety-critical scenarios.
We construct a traffic flow using $log2BT$, then vary the attributes of the participants nearby the ego gradually.
Specifically, $n$ varies from $\{3,4,6,10,12,16,20\}$. 
Then, we analyze the correlations among variables to find the main variables that cause critical events when $n=10$.



\subsection{Experiment Settings}
\noindent
All experiments are conducted in our Matrix simulator.
There are some parameters setting for Matrix-Fuzzer.
We set the threshold $\epsilon_{part}$ to 1.0, $\epsilon_{lat}$ to 2.0, and $\epsilon_{vel}$ to 1.0.
The coefficient $a$ of $minDist$ is -0.2, while the bias $b$ is 5.
The score weights are set as follows: $\alpha_1=1$, $\alpha_2=-1$, and $\alpha_3=0.2$.
All parameters are determined by several tests.

\subsection{Result Analysis: RQ1}
\begin{table}
	\centering 
	\caption{Trajectory Reconstruction}
	\label{tb:log2bt}
	\resizebox{0.98\linewidth}{!}{
		\begin{tabular}{c|c|c|c|c}
			\hline
			\hline
			\multirow{2}{*}{Partition} &
			\multicolumn{2}{c|}{NGSIM} &
			\multicolumn{2}{c}{ourData}\\
			
			\cline{2-5} 
			& $ADE_s$ (m) & $ADE_l$(m)& $ADE_s$ (m)& $ADE_l$ (m)\\
			\hline
			$log2BT_{log}$ & 0.57 & 0.21 & 0.48  & 0.16 \\
			$log2BT_{semantic}$ & 0.69 & 0.28 & 0.62  & 0.24 \\
			
			\hline
			\hline
		\end{tabular}
	}
\end{table}
\noindent
Fig. \ref{alg:log2bt} illustrates the performance of $log2BT$ in a selected trajectory.
Table \ref{tb:log2bt} shows the reconstruction errors for $log2BT$. 
There is no surprise that $log2BT_{log}$ performs better than $log2BT_{semantic}$ since some errors exist in classifying semantic behaviors, 
while $log2BT_{semantic}$ is more comprehensible.
$log2BT_{log}$ and $log2BT_{semantic}$ both achieve satisfactory reconstruction.
We see that the longitudinal error is less than 0.7m and the lateral error is less than 0.3m. 
It's acceptable as there is some safety distance between participants in NDE.
In particular, the origin trajectory may be covered when we vary the attributes of behaviors.
That indicates the availability and reasonability of $log2BT$ to generate $\mathcal{BT}$-based scenarios for latter tasks.

\subsection{Result Analysis: RQ2}

\begin{table}
	\centering 
	\caption{Efficiency Comparison}
	\label{tb:matrix_fuzzer}
	\resizebox{0.98\linewidth}{!}{
		\begin{tabular}{c|c|c|c||c|c|c}
			\hline
			\hline
			\multirow{2}{*}{Aspect} &
			\multicolumn{3}{c||}{Virtual cut in (Example \ref{exmp:cutin})} &
			\multicolumn{3}{c}{Real cut in (Example \ref{exmp:log_cutin})}\\
			
			\cline{2-7} 
			& Grid & AV-Fuzzer& Matrix-Fuzzer & Grid & AV-Fuzzer& Matrix-Fuzzer\\
			\hline
			$Total$ & 2240 & 840 &  320 & 2400 & 1231 & 360\\
			$Critical$ & 92 & 39 & 149  & 91 & 42 & 150 \\
			$Invalid$ & 1534 & 659 & 104  & 1183 &742 & 117\\
			$Types$ & 3 & 2 &  4 & 4 & 2 &4 \\
			\hline
			CR & 0.041 & 0.046 & \textbf{0.466}  & 0.038 & 0.034 & \textbf{0.435} \\
			IR & 0.685 & 0.785 & \textbf{0.325}  & 0.493 & 0.603 & \textbf{0.321} \\
			TR (\%) & 0.134 & 0.238 & \textbf{1.250}  & 0.167 & 0.162 & \textbf{0.111} \\
			\hline
			\hline
		\end{tabular}
	}
\end{table}
\noindent
Due to applying the early-stopping mechanism, the number of total scenarios generated by different methods may be different for the same $\mathcal{LS}$.
Table \ref{tb:matrix_fuzzer} summaries the simulation results for the competitors, while Figs. \ref{fig:search_process_1} and \ref{fig:search_process_2} illustrate the search process with iterations.
We have the following observations:
\begin{enumerate}
	\item The classic Grid search exhibits poor performance. It takes tremendous amount of cost and time, especially when the input space is larger.
	Grid search samples the parameters by pre-defined step shown in Figs. \ref{fig:grid_1} and \ref{fig:grid_2}. This method may find all types of safety-critical scenarios while the step is fine-grained, leading to unacceptable cost and lots of invalid scenarios.
	\item AV-Fuzzer finds many invalid critical scenarios. 
	The reason is that  
	it treats invalid collisions as safety-critical scenarios and
	misleads GA to an error direction. Thus, it generates next unreasonable scenarios iteratively shown in Figs. \ref{fig:av_fuzzer_1} and \ref{fig:av_fuzzer_2}.
	As a result, it only finds 2 types of real safety violations in Examples \ref{exmp:cutin} and Example \ref{exmp:log_cutin}. 
	\item Matrix-Fuzzer achieves the best performance in all aspects on both $\mathcal{LS}s$. 
	The method takes the reasonability of participant behaviors into account, 
	and enables to find the most types of violations with the fewest test scenarios (only around 30\% of that in AV-Fuzzer). 
	As shown as in Figs. \ref{fig:matrix_fuzzer_1} and \ref{fig:matrix_fuzzer_2}, it first covers the input space via adaptive random search and then applies BO to search locally.
	We observe that there are fewer invalid scenarios during BO search, which indicates the effectiveness of our method.
	Specifically, compared with AV-Fuzzer, it improves the CR and TR by 10x and 5.3x, respectively, while reduces the IR by 58.6\% in Example \ref{exmp:cutin}.
	In Example \ref{exmp:log_cutin}, it improves the CR and TR by 12.8x and 6.8x, respectively, while reduces the IR by 87.8\%.
\end{enumerate}


Different parameters lead to diverse types of the safety-critical scenarios.
As space limitation, we only analyze the types of safety-critical scenarios generated in the Example \ref{exmp:cutin}.
As shown in Fig. \ref{fig:matrix_fuzzer_1}, $S_1$, $S_2$, and $v$ are more likely to cause the violations as they converge to ranges.
To demonstrate the types of safety-critical scenarios, we visualize the search space in 3D shown in Fig. \ref{exp:3d}.
We first cluster the parameters into 4 groups and each center represents a type of violation.
In addition, we analyze the clusters to find root causes. 
Fig. \ref{exp:cutin_type} illustrates the invalid type and the corresponding 4 types of violations.
We observe that Type 1 is caused by a small $S_1$ while Type 2 is due to the $S_2$. 
Both of them leave few time for the ego to react, thus resulting in collisions.
Types 3 and 4 are main caused by the lower end speed $v$ of \textsf{changelane} behavior, but with different $S_2$.
These root causes help us to sample representative parameters from the simulation testing to on-road testing.

\subsection{Result Analysis: RQ3}

\begin{table}
	\centering 
	\caption{Search Algorithm Comparison}
	\label{tb:matrix_fuzzer_hd}
	\resizebox{0.98\linewidth}{!}{
		\begin{tabular}{c|c|c|c|c|c|c|c}
			\hline
			\hline
			\multirow{2}{*}{$n$} &
			\multicolumn{7}{c}{BO/GA} \\
			
			\cline{2-8} 
			 & $Total$ & $Critical$& $Invalid$ & $Types$ & CR & IR & TR (\%)\\
			\hline
			3 & 210/428 & 63/32 & 96/320  & 3/3 & 0.300/0.075 & 0.457/0.748 & 1.429/0.701 \\
			4 & 310/550 & 94/74 & 149/417 & 4/3 & 0.303/0.134 & 0.481/0.758 & 1.290/0.545\\
			6 & 450/827 & 171/235 & 211/433  & 5/5 & 0.380/0.284 & 0.469/0.524 & 1.111/0.600 \\
			\hline
			\textbf{10} & 744/1282 & 247/394 & 357/712  & 8/8 & 0.332/0.307 & 0.479/0.555 & 1.08/0.624 \\
			\hline
			12 & 850/1071 & 194/249 & 467/549  & 10/10 & 0.228/0.232 & 0.549/0.512 & 1.176/0.934 \\
			16 & 1483/1171 & 307/249 & 889/645  & 10/10 & 0.207/0.212 & 0.599/0.551 & 0.674/0.854 \\
			20 & 1890/1373 & 348/432 & 1125/744  & 9/10 & 0.184/0.314 & 0.595/0.542 & 0.476/0.728 \\
			\hline
			\hline
		\end{tabular}
	}
\end{table}

\noindent
Table \ref{tb:matrix_fuzzer_hd} summaries the results of different search algorithms varying the number of variables $n$.
As $n$ grows, the input space expands exponentially, making it much hard to find the safety-critical scenarios.
We observe that BO performs better than GA when $n$ is small, while GA has better performance as $n$ grows.
Therefore, we default BO as the search algorithm in our fuzzing engine when $n<10$. Otherwise, GA is selected.

To analyze the weights of variables contributing to the safety-critical scenarios, we visualize the search process, search space, and correlation matrix among variables.
The divergences of variables indicate their respective weights. Specifically, the weight becomes smaller as its divergence increases.
This inspires us to focus on varying the main variables and pay less attention to the inessential variables.
As show in Figs. \ref{exp:hd}, the variables $a_1-s_1$, $a_1-v_2$ and $a_2-s_1$ may have larger weights since they are convergent.
Fig. \ref{exp:hd_3d} illustrates their search space to locate the critical space causing the violations.
In addition, we mine the correlations among variables shown in Fig. \ref{exp:hd_correlation}.
We observe that variables $a_1-s_1$, $a_1-v_2$ and $a_2-s_1$ may have inherent relationships, 
which inspires us to construct conditional safety-critical scenarios.
We leave it to the further work.

\section{Conclusion}\label{sec:conclusion}
\noindent
In this paper, we propose the Matrix-Fuzzer, a behavior tree-based simulation testing framework, to search safety-critical scenarios for autonomous driving systems.
In overall, we propose the $log2BT$ method to abstract logged trajectories to behavior trees, set the variable space from the real-world driving distributions, and design an adequate evaluation engine to guide the adaptive algorithm generating valid test scenarios.
Our approach is evaluated on our Matrix simulator.
The experimental results show that our $log2BT$ has satisfactory longitudinal and lateral errors in trajectory reconstruction. 
Our approach is able to search the most types of safety-critical scenarios, but only generating around 30\% of the total scenarios compared with the baseline algorithms.
Specifically, it improves the ratio of the critical violations to total scenarios and the ratio of the types to total scenarios
by at least 10x and 5x, respectively, while reducing the ratio of the invalid scenarios to total scenarios by at least 58\% in two case studies.

\bibliographystyle{abbrv}
\bibliography{myRef}


\end{document}